\documentclass[conference]{IEEEtran}
\IEEEoverridecommandlockouts
\usepackage{graphicx}
\usepackage[ruled]{algorithm2e}
\usepackage{algorithmic}
\usepackage{multirow}
\usepackage{amsmath} 
\usepackage{array}
\usepackage{arydshln}
\usepackage{dsfont}
\usepackage{amsthm}
\usepackage{amsfonts}
\usepackage{booktabs}
\usepackage{setspace}
\usepackage{subcaption}
\usepackage{mathtools}
\usepackage{hyperref}
\usepackage{xcolor}
\DeclareMathOperator*{\argmax}{arg\,max}

\begin{document}
\title{Distribution-Based Trajectory Clustering}
%
\author{
\IEEEauthorblockN{
Zi Jing Wang\IEEEauthorrefmark{1}, 
Ye Zhu\IEEEauthorrefmark{2}, 
Kai Ming Ting\IEEEauthorrefmark{1}}
\IEEEauthorblockA{ \IEEEauthorrefmark{1}National Key Laboratory for Novel Software Technology, Nanjing University, China}
\IEEEauthorblockA{\IEEEauthorrefmark{2} Centre for Cyber Resilience and Trust, Deakin University, Australia}
\IEEEauthorblockA{Email: wangzj@lamda.nju.edu.cn, ye.zhu@ieee.org, tingkm@nju.edu.cn}
}
\maketitle              
\begin{abstract}
Trajectory clustering enables the discovery of common patterns in trajectory data. Current methods of trajectory clustering rely on a distance measure between two points in order to measure the dissimilarity between two trajectories. The distance measures employed have two challenges: high computational cost and low fidelity. Independent of the distance measure employed, existing clustering algorithms have another challenge: either effectiveness issues or high time complexity. In this paper, we propose to use a recent Isolation Distributional Kernel (IDK) as the main tool to meet all three challenges. The new IDK-based clustering algorithm, called TIDKC, makes full use of the distributional kernel for trajectory similarity measuring and clustering. TIDKC identifies non-linearly separable clusters with irregular shapes and varied densities in linear time. It does not rely on random initialisation and is robust to outliers. An extensive evaluation on 7 large real-world trajectory datasets confirms that IDK is more effective in capturing complex structures in trajectories than traditional and deep learning-based distance measures. Furthermore, the proposed TIDKC has superior clustering performance and efficiency to existing trajectory clustering algorithms. 
\end{abstract}

\section{Introduction}


Clustering trajectories is an important task in trajectory data mining, which aims to compute the similarity among trajectories in a given dataset, and 
group trajectories in a given dataset into clusters of similar trajectories. 
These trajectory clusters provide invaluable information in relation to individual objects' movements. This knowledge is often used in further decision-making. Trajectory clustering is widely used in traffic monitoring  \cite{StefanAtev2006LearningTP}, path prediction  \cite{ZaibenChen2011DiscoveringPR} and movement behaviour  analysis  \cite{ShinAndo2011RoleBehaviorAF}. 

Trajectory clustering is a non-trivial problem, and it has three key challenges:
\begin{enumerate}
        \item \textbf{Existing trajectory distance measures have high time complexity} because their core computations rely on a point-to-point distance measure. To compute the distance of two trajectories, each having $m$ points, these distance measures have at least $O(m^2)$ time cost. 
    \item \textbf{Low fidelity of existing distance measures}. Unlike tabular data, the meaning of distance between trajectories is not straightforward because a trajectory consists of  sampled records of  spatial
and temporal attributes in real time. 
    Most existing distance measures are sensitive to sampling rate, outliers, spatial/temporal lengthening and contraction \cite{HanSu2020ASO}.
    \item \textbf{Existing trajectory clustering algorithms have either effectiveness issues or high time complexity}. For example, $k$-means-like clustering algorithms \cite{CORDEIRODEAMORIM20121061} are unable to find arbitrarily shaped clusters; while spectral clustering algorithm \cite{gao2020ship} has worse than linear time complexity.  
\end{enumerate}

No good solutions have been offered in the literature to meet all the above challenges. Although recent deep learning-based representations and clustering methods show potential improvements, we found that they have yet to produce a satisfactory  outcome (see the experimental section for details).

Here we propose to use a recent Isolation Distributional Kernel (IDK)  \cite{KaiMingTing2020IsolationDK} as the main tool to meet all the three challenges.
IDK is used in the first instance to meet the first two challenges as a representation-cum-measure: a trajectory in $\mathbb{R}^d$ is mapped into a new representation as a point (vector) in the feature space of IDK; and the similarity between two trajectories can then be computed as a dot product of two mapped points in the feature space. 

While it is possible to use any existing clustering algorithms on the set of points in the feature space, we find that their clustering outcomes are less than desirable. We propose a new IDK-based clustering algorithm, called TIDKC, to meet the third challenge\footnote{Code is at https://github.com/IsolationKernel/TIDKC}. TIDKC is unique because has all the following characteristics:
\begin{enumerate}
    \item Able to identify non-linearly separable clusters with irregular shapes and varied densities.
    \item Run in linear time.
    \item Does not rely on random initialisation and is not sensitive to outliers.
\end{enumerate}

\begin{table*}[t]
\caption{Different similarity measures for trajectory data. $\mathcal{P}$ denotes a distribution based representation where the distribution is modelled or requires no modelling.}
\small
\begin{tabular}{@{}ccc@{}cc@{}c@{}}
\hline
 {Category}    &  {Distance/measure}  & Focus  &  {Robust to noise} &  {Time complexity}         & {Examples} \\ 
\hline
\multirow{3}{*}{Point based} & Hausdorff Distance \cite{AbdelAzizTaha2015AnEA} & {point set} & $\times$               & $O(n^2)$  &   CTHD \cite{chen2011clustering}  \\
    & DTW Distance \cite{CMyers1980PerformanceTI}       & {time wrapping} & $\checkmark$  & $O(n^2)$  &    {ERP \cite{chen2005robust}, PDTW \cite{keogh2000scaling}  }    \\
    & Fréchet Distance \cite{AJWard1954AGO}   & {point set} & $\times$  & $O(n^2\log n^2)$    & Discrete Fréchet  \cite{eiter1994computing} \\
    \hline
\multirow{2}{*}{Distribution based} & Earth Mover's  Distance  \cite{rubner1998metric}  & {$\mathcal{P}$ is modeled as histogram}  & {$\checkmark$} & {$O(n^3)$} &     {DBF \cite{yang2022fast}}     \\
    & IDK  \cite{KaiMingTing2020IsolationDK}   & {$\mathcal{P}$ requires no modeling} & $\checkmark$    & $O(n)$   &   Anomaly detection \cite{wang2023principled}       \\
\hline
\label{tab:similarity}
\end{tabular}
\end{table*}

The contributions of this paper are :
\begin{enumerate}
  \item  Proposing to use the distributional kernel IDK to represent trajectories and measure the similarity between two trajectories for clustering.  IDK has high precision in retrieving the most similar trajectories because they are ranked based on similarity in distributions. 
     \item  Developing a new clustering algorithm TIDKC that makes full use of the distributional kernel for trajectory clustering. It is the first clustering algorithm that employs two levels of the distributional kernel to perform trajectory clustering with linear time complexity. 
     \item Verifying the effectiveness and efficiency of the proposed IDK and TIDKC  on $7$ large real-world trajectory datasets. Our empirical results confirm that (a) IDK is more effective in capturing complex structures in trajectories, and it is more effective in computing similarity between trajectories than existing distance measures including the deep learning-based t2vec; and (b) TIDKC produces better trajectory clustering outcomes than existing methods including a deep learning clustering; and it runs faster than existing trajectory clustering algorithms such as K-Medoids and spectral clustering.  
\end{enumerate}


The rest of the paper is organised as follows. Section~\ref{rela} provides related works on  trajectory distance measures and clustering. Section \ref{ker} and Section \ref{TIDKC} introduce IDK and TIDKC for trajectory distance measuring and clustering, respectively. Section~\ref{TGDKC} describes a variant of TIDKC. Section~\ref{Eva} presents an empirical evaluation, followed by a conclusion in Section~\ref{sec-conclusions}. 

\section{Related work}
\label{rela}
We survey methods for trajectory distance measures and trajectory clustering in the following two subsections.

\subsection{Trajectory Distance Measures}


\textit{Definition 1}: A trajectory $T$ of a travelling  object consists of a sequence of location points $x_i \in \mathbb{R}^d$, where $i$ denotes the time/order the object has travelled: $T = [x_{1}, x_{2}, \cdots, x_{n}] $.

Trajectory distance measurement computes the distance/dissimilarity between two trajectories. The larger the distance is, the less similar the two trajectories are. 
We categorise existing measures into three categories, point-based, distribution-based and deep learning-based methods as follows. 
Table \ref{tab:similarity} shows a summary of different similarity measurements on trajectory data and their characteristics.

\paragraph{Point-based measures}
There are two subcategories of point-based measures. The first subcategory is the conventional measures such as Hausdorff distance and (Discrete) Fréchet distance. These distance measures operate on sets, where each trajectory is assumed to be a set of points $x \in \mathbb{R}^d$. For example, by treating each trajectory as a set, Hausdorff distance between two trajectories $T_1$ and $T_2$ is defined as: 
\begin{align}
    d_{H}(T_1, T_2) = \max \left(\sup_{x\in T_1} d_{min}(x, T_2), \sup_{y\in T_2} d_{min}(y, T_1)\right),
\end{align}
where $d_{min}(x, T)$ denotes the minimum distance of point $x$ to all points in $T$. 

Note that Hausdorff distance (so as Fréchet distance) effectively computes the final distance between $T_1$ and $T_2$ based on \emph{two critical points} $\hat{x} \in T_1$ and $\hat{y} \in T_2$ only, i.e., 
$d_{H}(T_1, T_2) \equiv\ \parallel \hat{x} - \hat{y} \parallel$. 
This applies to (Discrete) Fréchet distance too. 
Hausdorff Distance is able to handle trajectories with non-unified length, while its time complexity runs to $O(mn)$, where $m,n$ is the length of each trajectory.

The second subcategory is based on Dynamic Time Warping (DTW) distance, which is a commonly used measure in sequences \cite{CMyers1980PerformanceTI}. It aims at finding the optimal matching of two trajectories by exploiting temporal distortions on trajectories. After finding the best point-pairs alignment of two trajectories, it computes the sum of  Euclidean distances of all best-matched point-pairs.

%
All the distance measures above have at least quadratic time complexity.

\paragraph{Distribution-based measure 1}

Earth Mover’s Distance (EMD) assumes that one trajectory is a set of identical and independent distributed (i.i.d.) points, sampled from a probability distribution $\mathcal{P}$.  EMD \cite{rubner1998metric,yang2022fast} employs a histogram to model a distribution. For two trajectories modelled as probability distributions $\mathcal{P}$ and $\mathcal{Q}$, EMD aims at finding an optimal transportation plan from $\mathcal{P}$ to $\mathcal{Q}$ that minimizes the overall cost:
\begin{align}
\operatorname{EMD}(\mathcal{P}, \mathcal{Q})=\inf _{\gamma \in \Pi(\mathcal{P}, \mathcal{Q})} \mathbb{E}_{(x, y) \sim \gamma}[d(x, y)],
\end{align}
\noindent
where $\Pi(\mathcal{P}, \mathcal{Q})$ is the set of all joint distributions whose marginals are $\mathcal{P}$ and $\mathcal{Q}$.


\paragraph{Distribution-based measure 2: Isolation Distributional Kernel (IDK)}
Current similarity measures used in existing trajectory clustering usually have high computational complexity and are sensitive to local fluctuation/noise. 
In this paper, we propose to use the distributional kernel IDK \cite{KaiMingTing2020IsolationDK} to represent trajectories and measure the similarity between two trajectories. 
Unlike existing distance measures mentioned earlier which have at least quadratic time complexity,
IDK has linear time complexity. Moreover, IDK has unique data-dependent  and  injective properties (see below). IDK has shown good performance in point anomaly detection tasks  \cite{KaiMingTing2020IsolationDK}. This has inspired us to apply IDK to trajectories. They are the key factors that lead to improvement in accuracy over existing distance measurements in the anomalous trajectory detection task \cite{wang2023principled}. However, how to take full advantage of those factors of IDK to cluster trajectories effectively is still unclear.

IDK is  based on kernel mean embedding \cite{KrikamolMuandet2016KernelME} which extends a point-to-point kernel $\kappa$ to measure the similarity between two distributions. It maps a distribution $\mathcal{P}$ as a point in a reproducing kernel Hilbert space (RKHS)  \cite{KrikamolMuandet2016KernelME} via its feature map $\widehat{\Phi}$, representing each distribution $\mathcal{P}$ on data support $\mathcal{X}$ as a mean function:
\begin{align}
    \widehat{\Phi}(\mathcal{P})=\int_{\mathcal{X}} \kappa(x, \cdot) \mathrm{d} \mathcal{P},
\end{align}
where $\kappa$ is a symmetric and positive definite kernel function. 


Typically, a Gaussian kernel is used as $\kappa$, producing Gaussian distributional kernel (GDK), and the estimated similarity between two distributions $\mathcal{P}_{S}$ and $\mathcal{P}_{T}$, which generate sets $S$ and $T$ respectively, is computed as follows \cite{KrikamolMuandet2016KernelME}:
\begin{align}
    \mathcal{K}_G (\mathcal{P}_{S}, \mathcal{P}_{T}) = \frac{1}{|S||T|} \sum_{x \in S, y \in T} \kappa_G(x, y).
    \label{IDKmearue}
\end{align}


Instead of Gaussian kernel, Isolation Kernel (IK) \cite{KaiMingTing2018IsolationKA} has been used in kernel mean embedding to produce Isolation Distributional Kernel  \cite{KaiMingTing2020IsolationDK} (IDK). 

Unlike a typical point-to-point kernel which has a closed form expression, IK, denoted as $\kappa_I$, has none but derives its feature map $\phi$ directly from a given dataset $D$, i.e.,
\[\kappa_I(x,y \mid D) = \left\langle \phi(x \mid D), \phi(y \mid D) \right\rangle.
\]
Two distinguishing aspects of IK \cite{KaiMingTing2018IsolationKA} are:
\begin{enumerate}
    \item Data-dependent similarity: two points in a sparse region are more similar than two points of equal inter-point distance in a dense region.
    \item Its finite-dimensional feature map is derived from a dataset. Most commonly used kernels have a feature map which has an intractable dimensionality.
\end{enumerate}


The empirical estimation of IDK for two distributions is provided as follows:
\begin{align}
    {\mathcal{K}}_{I}\left(\mathcal{P}_{S}, \mathcal{P}_{T} \mid D\right) &=\frac{1}{|S||T|} \sum_{x \in S,y \in T} \left\langle \phi(x \mid D), \phi(y \mid D) \right\rangle \nonumber \\  
    &=\left\langle\widehat{\Phi}\left(\mathcal{P}_{S} \mid D\right), \widehat{\Phi}\left(\mathcal{P}_{T} \mid D\right)\right\rangle,
    \label{eqn_IDK}
\end{align}
where $\widehat{\Phi}\left(\mathcal{P}_{S} \mid D\right)=\frac{1}{|S|} \sum_{x \in S} \phi(x \mid D)$ is the kernel mean map of IDK. 


Though both are distributional measures, IDK has three advantages over EMD in measuring the similarity of two trajectories. First, IDK computes the similarity without explicitly modelling a distribution. EMD must model a distribution, for example, as a histogram. This modelling could introduce errors even before the similarity is computed. Second, the kernel mean map is injective\footnote{Only if the kernel $\kappa$ employed is a characteristic kernel \cite{KrikamolMuandet2016KernelME}. Isolation Kernel has been shown to be a characteristic kernel \cite{KaiMingTing2020IsolationDK} (so as Gaussian kernel).}, {\it i.e.}, $\Vert \widehat{\Phi}(\mathcal{P}_S) - \widehat{\Phi}(\mathcal{P}_T) \Vert = 0$ if and only if $\mathcal{P}_S = \mathcal{P}_T$ \cite{KaiMingTing2020IsolationDK}. This means that \emph{similar trajectories are not to be ranked lower than dissimilar trajectories in search of the most similar trajectories}. EMD has yet to be shown to have such a property. In fact, \emph{all existing measures including representation learning methods  have no such property}. The exceptions are IDK and GDK. Third, IDK has linear time complexity because the mapping of each trajectory of length $m$ costs $O(m)$, and the actual similarity computation of Equation~(\ref{eqn_IDK}) using the dot product has constant time. 
In contrast, EMD has at least quadratic time complexity.   

The intuition is to treat each trajectory as a sample from an unknown multi-dimensional probability density function (pdf). With this treatment, any distributional measures for unknown multi-dimensional pdf, such as IDK and GDK, can be used to measure the similarity of two trajectories. 

\paragraph{Deep learning-based methods}
A deep neural network method, t2vec, learns a representation for trajectories  \cite{XiuchengLi2018DeepRL}. It uses a sequence encoder-decoder model to learn from trajectories and generate a vector representation for each trajectory. This method is robust to non-uniform and noisy trajectory data and is able to compute the similarity of two trajectories in linear time.

NeuTraj \cite{yao2019computing} trades off accuracy with efficiency by learning a distance function that approximates a given distance function such as DTW distance or Hausdorff distance.
It reduces the time complexity of the original distance function from $O(n^2)$ to $O(n)$ where $n$ is the length
of each of the two trajectories under measurement. TrajGAT \cite{yao2022trajgat} is a recent improvement over NeuTraj.




\subsection{Clustering Methods}
Trajectory clustering algorithms can be generally divided into partition-based and density-based methods \cite{wang2021survey}. 


\begin{figure*}[!t]
\centering
	\begin{subfigure}{0.3 \textwidth}
		\centering
		\includegraphics[width=0.7\textwidth]{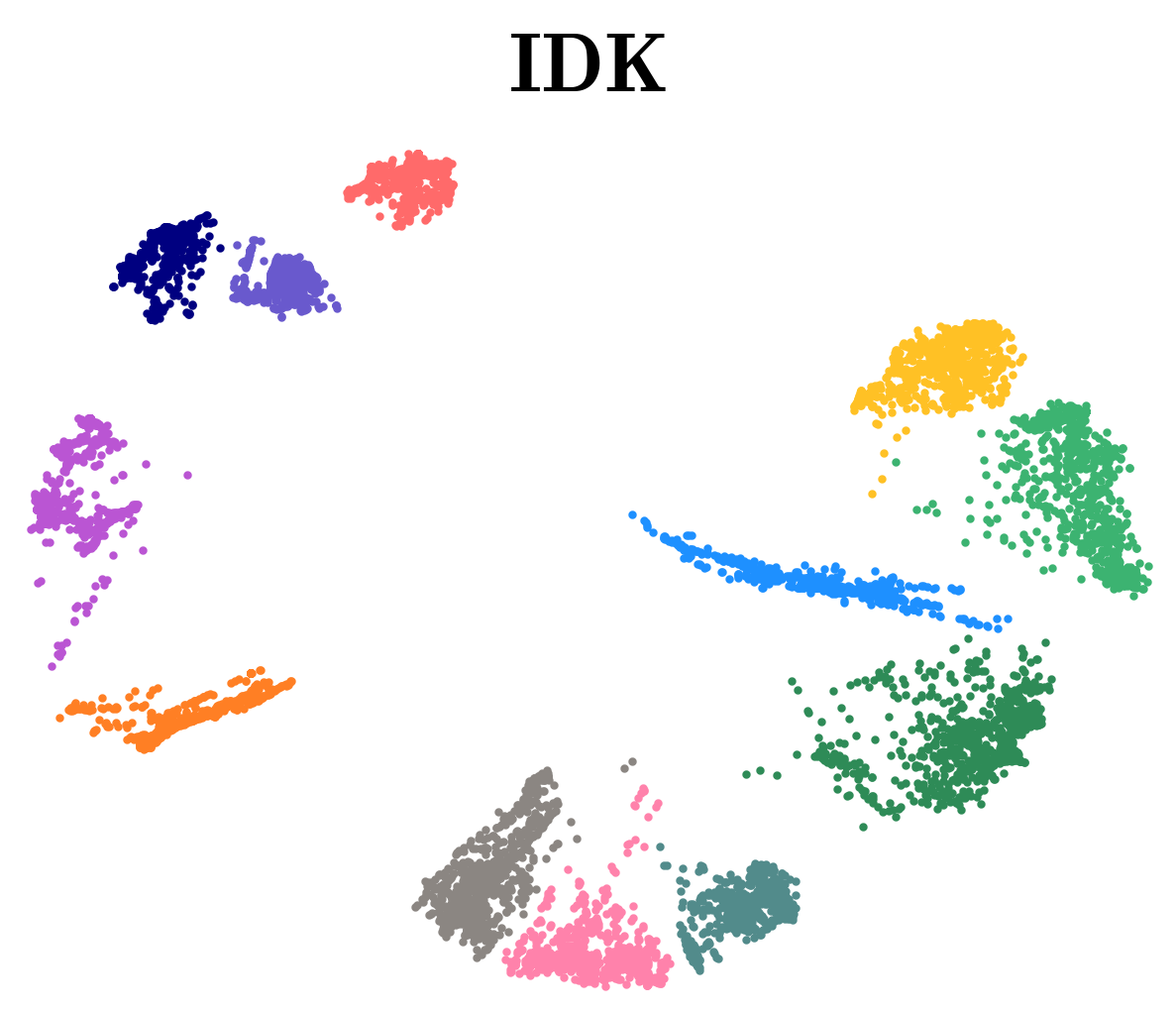}
	\end{subfigure} 
	\begin{subfigure}{0.3 \textwidth}
		\centering
		\includegraphics[width=0.7\textwidth]{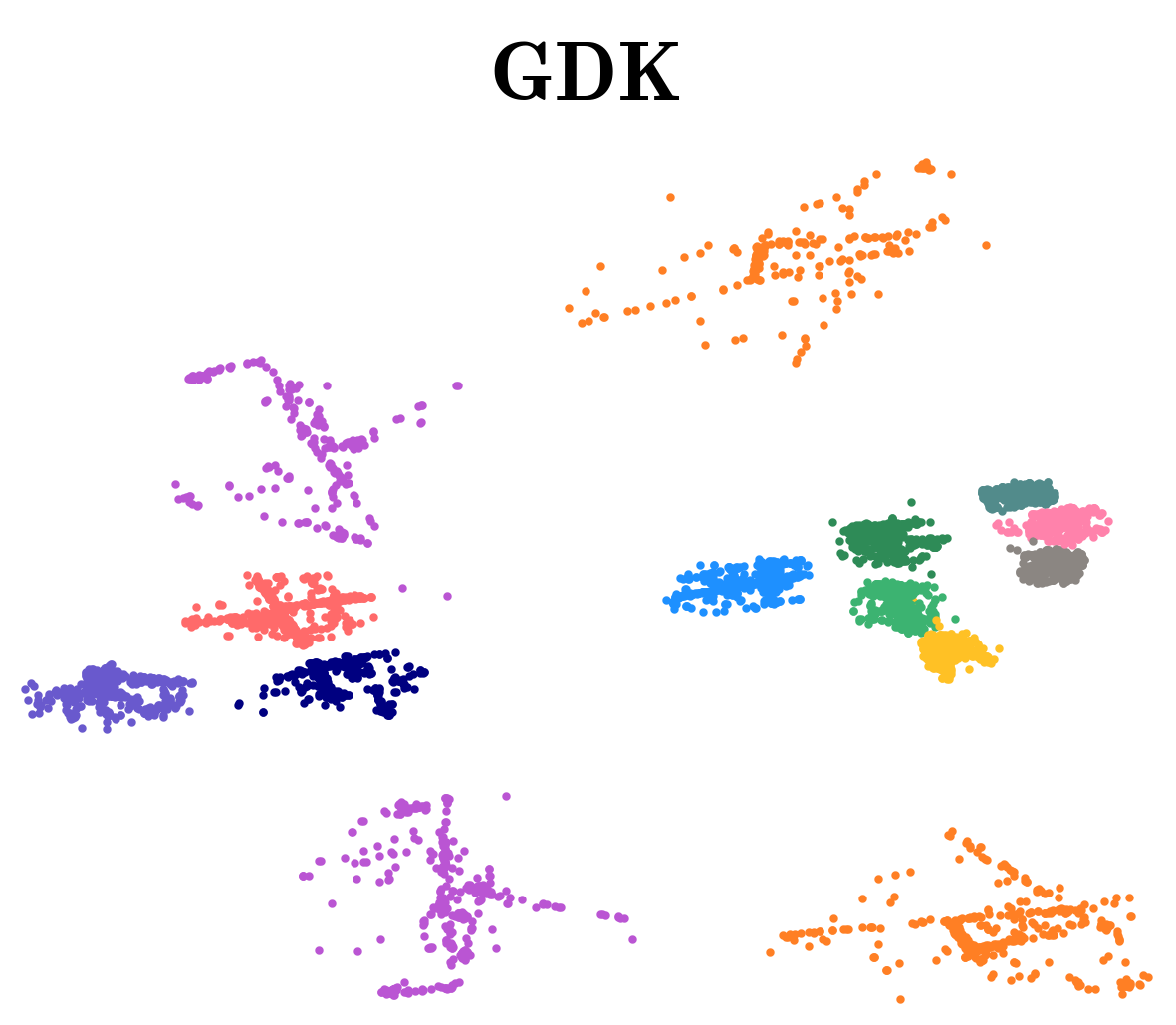}
	\end{subfigure}\\
	\begin{subfigure}{0.3 \textwidth}
		\centering
		\includegraphics[width=0.7\textwidth]{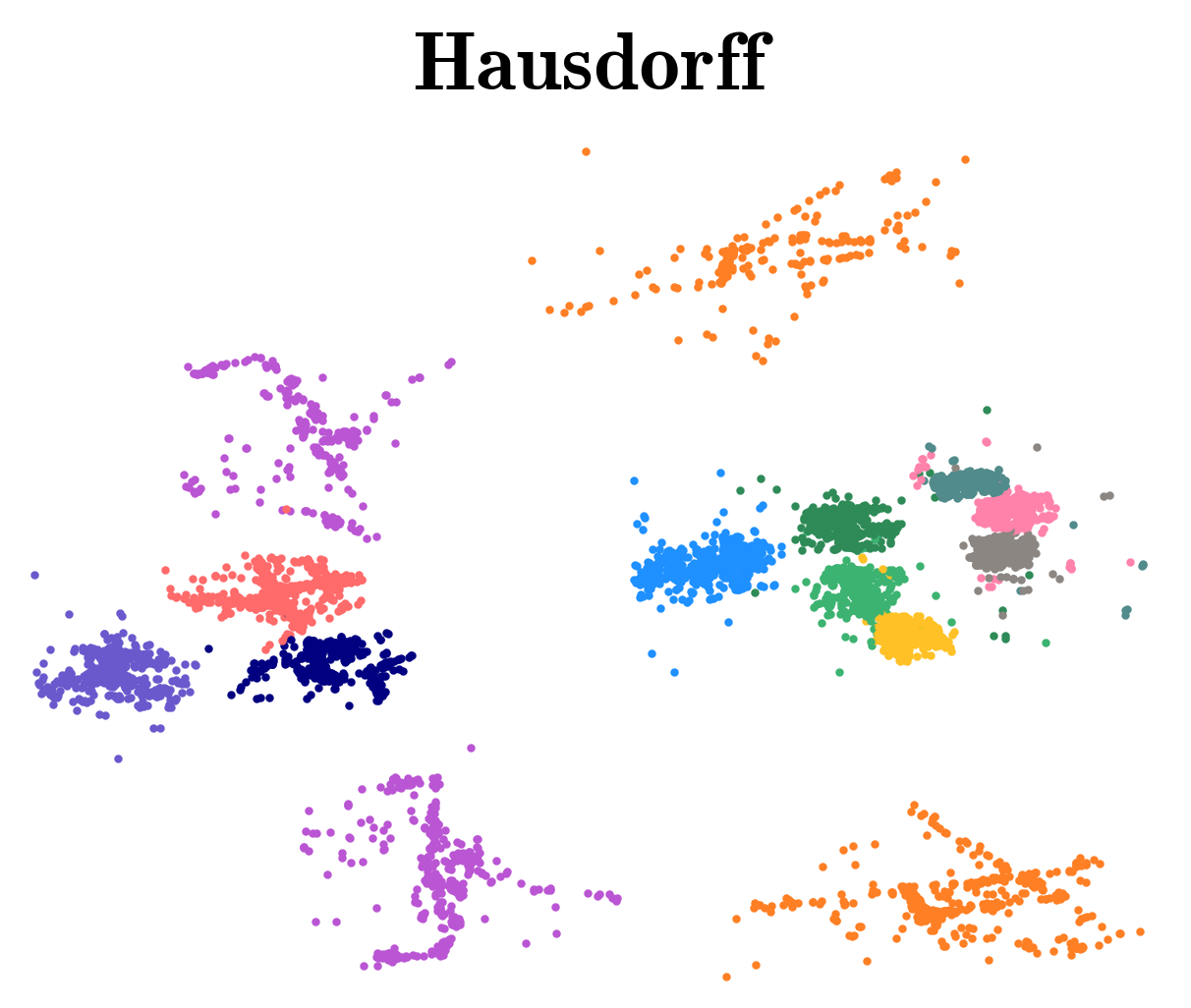}
	\end{subfigure} 
	\begin{subfigure}{0.3 \textwidth}
		\centering
		\includegraphics[width=0.7\textwidth]{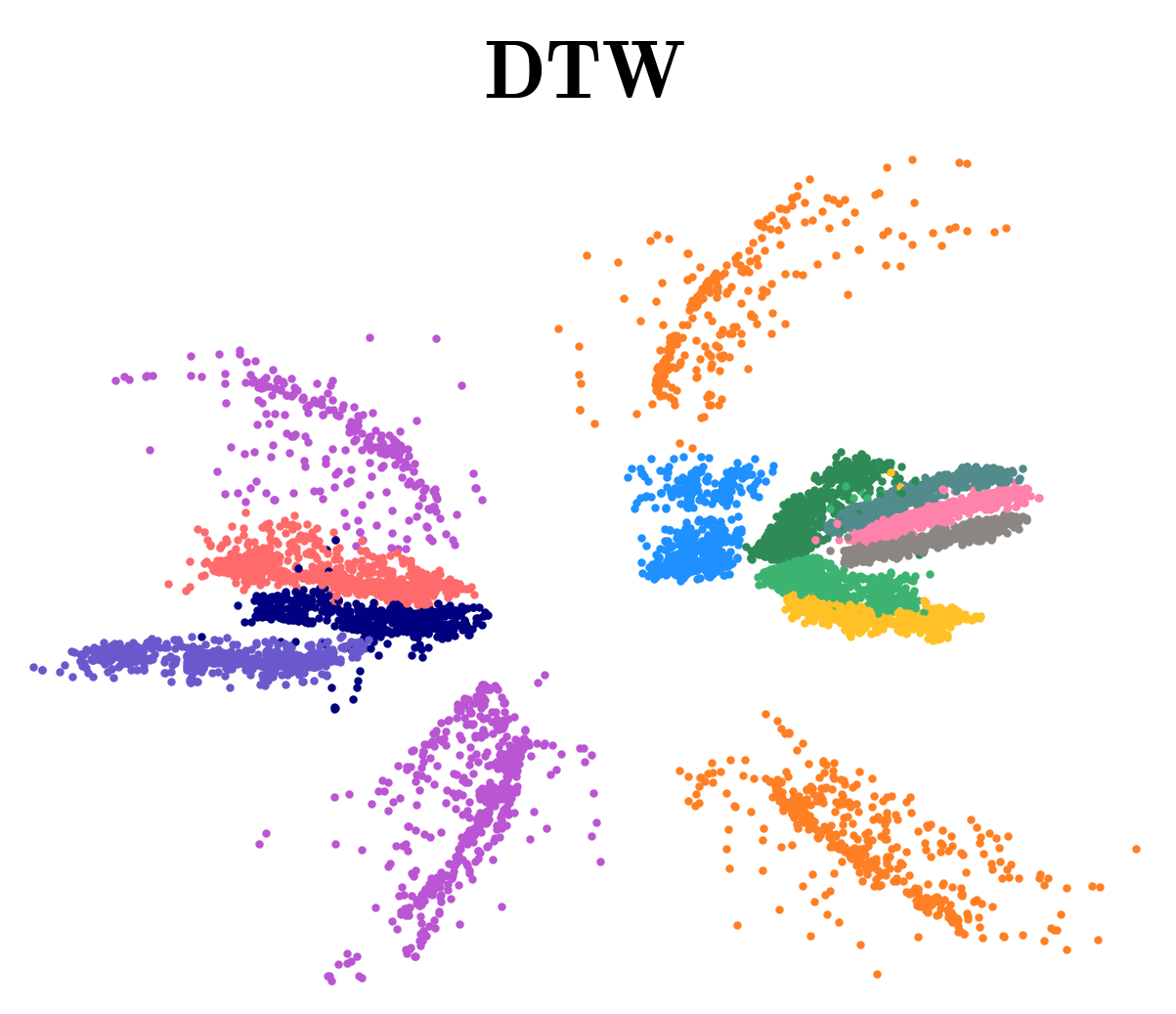}
	\end{subfigure} 
	\begin{subfigure}{0.3 \textwidth}
		\centering
		\includegraphics[width=0.7\textwidth]{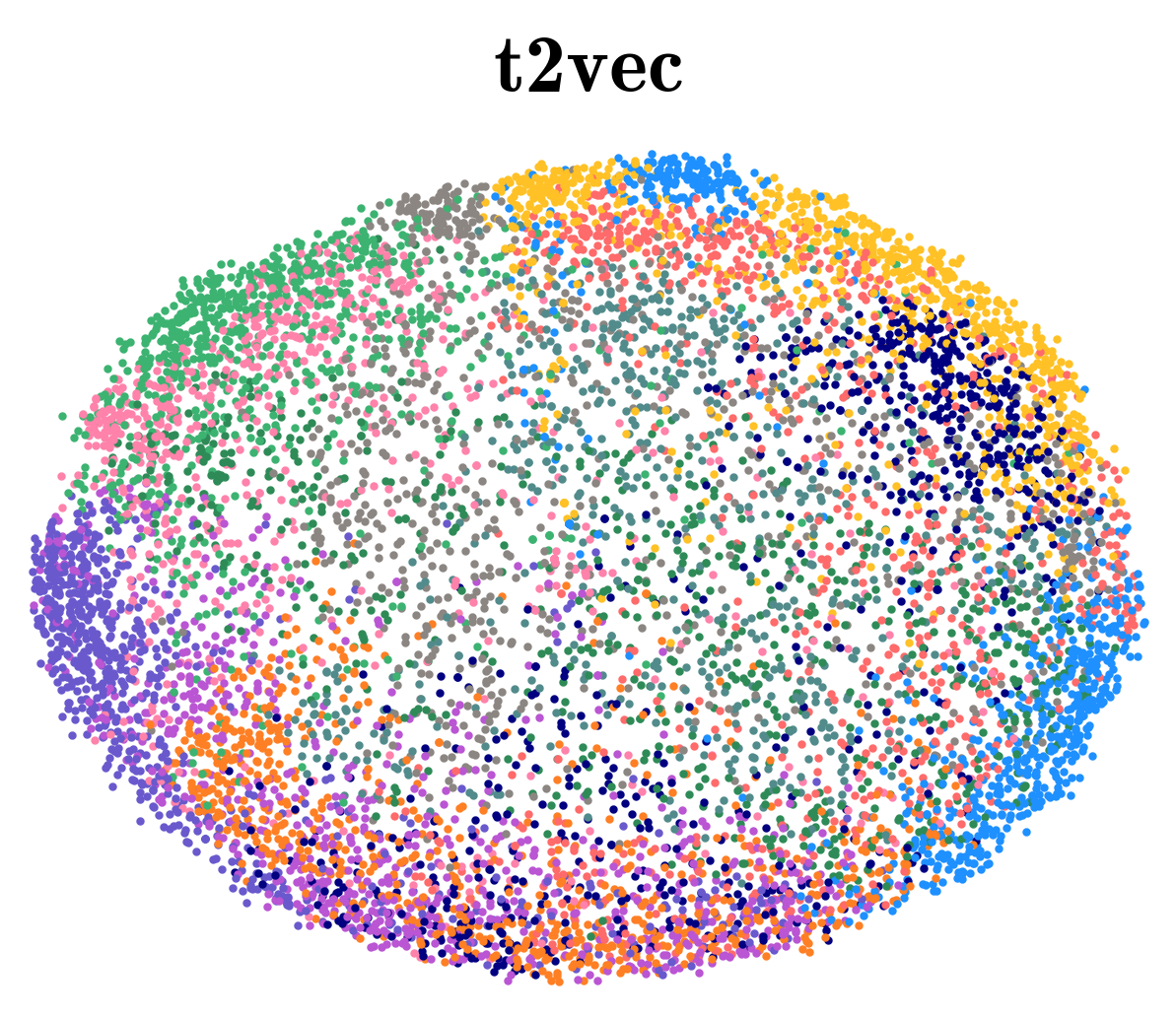}
	\end{subfigure}
	\caption{MDS visualization of relative distances between points on the Geolife dataset, as measured using five distance measures. Note that each of the purple and orange clusters is split into two sub-clusters in GDK, HD and DTW.}
	\label{fig:mds}
\end{figure*}

\paragraph{Partition-based methods}
Partition-based methods divide trajectories into  $k$ non-overlapping clusters. Many extensions of K-Means and K-Medoids are proposed to deal with trajectory data clustering. Those clustering methods are usually computationally intractable for large trajectory datasets; thus, they often must reduce the number of trajectories or through
some approximation \cite{wang2021survey}. 
For example, Xu et al. \cite{xu2015unsupervised} propose a multi-kernel-based estimation process to reduce the total number of trajectories  to a smaller number of shrunk transformed trajectories. Then traditional k-means with Euclidean distance can be applied on the shrunk transformed trajectories. Another clustering 
k-paths  \cite{ShengWang2019FastLT}  aims to identify $k$
``representative'' paths in a road network. It relies on a quasi-linear edge-based-distance (EBD) measure and prunes the search space to deal with large-scale trajectory data. The EBD is calculated based on the intersecting road segments and travel length. 
To reduce the number of distance computations, k-paths makes use of indexing and pruning techniques. Different from K-Means based clustering that can only detect globular clusters, spectral clustering has the ability to handle clusters with arbitrary shapes \cite{gao2020ship}. The key problem for spectral-based trajectory clustering is to construct the affinity graph that represents the pairwise relations between trajectories \cite{ChihChiehHung2015ClusteringAA}. 
To obtain additional features for more reliable segmentation, Liu et al. \cite{liu2005motion} model the high-order motion for similarity calculation before applying spectral clustering.
Liu et al. \cite{liu2005motion} model motion vectors for each trajectory via segmentation; and
use a normalised correlation between the motion vectors of trajectories as a similarity measure, before using a spectral clustering algorithm \cite{shi1998motion}. 
ReD-SFA \cite{zhang2016red} learns the trajectory features and constructs a graph 
simultaneously as a problem of cross entropy minimisation by reducing the discrepancy between two Bernoulli distributions parameterized by the updated distances and the affinity graph, respectively. Then it conducts spectral clustering based on the final affinity graph. However, spectral clustering has fundamental limitations, e.g., high computational complexity and difficulty in identifying clusters with different densities  \cite{BoazNadler2006FundamentalLO}.


\paragraph{Density-based methods}

Density-based algorithms usually identify dense segments and then connect these segments to form representative routes. This kind of algorithm discovers clusters with arbitrary sizes and shapes. For example, TRACLUS \cite{lee2007trajectory} aims to discover common sub-trajectories from a set of trajectories.  It splits each trajectory into line segments and then groups dense line segments into clusters based on DBSCAN \cite{ester1996density}. The distance between two line segments is calculated based on three components, including perpendicular distance, parallel distance and angle distance. Nevertheless, existing density-based clustering algorithms trade off accuracy and efficiency for clustering massive trajectories. Extracting exact sub-trajectories used to represent each cluster is an NP-hard problem \cite{wang2021survey}.

\paragraph{Deep Learning-based Clustering}

Deep learning is attracting increasing interest and has made some progress for trajectory clustering in the last few years. There are new trends in trajectory generation, trajectory representation learning and similarity measurement \cite{wang2021survey}.
E$^2$DTC  \cite{ZiquanFang2021E2D} is an end-to-end deep trajectory clustering framework that requires no manual feature extractions. It employs an encoder-decoder scheme to learn the representation of trajectories, and then uses a self-training method to jointly learn the cluster-oriented representation and perform the clustering analysis. Another work  \cite{TobiasSchreck2008VisualCA} applies a self-organising map (SOM) on trajectory data to build a visual monitoring and interaction system for trajectory clustering. Deep transport  \cite{XuanSong2016DeeptransportPA} employs a long short-term memory network to simulate human mobility mode at a city wise level, and predicts people's transportation pattern based on their trajectories.

\section{Measuring trajectory distance using IDK}
\label{ker}
In this section, we investigate the key properties of IDK and other trajectory similarity measurements. 




Conventional measures like Hausdorff distance and Fréchet distance treat trajectories as sets of points and compute  the distance based on two critical points of two trajectories. Instead, we treat a trajectory as a set of independent and identical distributed (i.i.d.) points sampled from an unknown distribution, which is similar to the assumptions made by EMD for trajectories. The distance between two distributions is used to measure the distance between trajectories. Unlike EMD, we do not use histograms to model the trajectory distribution, but compute the distance between distributions directly via IDK. The additional distributional information provides a useful means for trajectory representation and similarity measurement. 

A clustering method requires a distance measure to group similar objects together. A good distance measure should produce a short distance to the neighbours of an object from the same cluster. We conduct MDS\footnote{Multidimensional scaling (MDS)~ \cite{borg2012applied} performs a transformation from a high-dimensional space to a $2$-dimensional space for visualisation by preserving as well as possible the pairwise global distances in both spaces.} visualisation and retrieval evaluation on a large Geolife dataset\cite{ZiquanFang2021E2D} to show the effectiveness of IDK as a trajectory measure in trajectory clustering tasks. 

\paragraph{MDS Visualisation}
Figure \ref{fig:mds} shows the results  using five distance measures. The IDK measurements congregate mapped points of trajectories of every cluster well, i.e., there are clear gaps between different clusters.  In contrast, each of GDK, Hausdorff distance and DTW distance splits some sparse clusters (e.g., the purple and orange clusters shown in Figure \ref{fig:mds}) into two sub-clusters; Hausdorff distance produces quite a few scatter points; and the gaps between clusters are too close with DTW. t2vec has the worst result as there is no clear gap between any two clusters. 

\begin{figure}[!tb]
    \centering
    \includegraphics[width=0.95\linewidth]{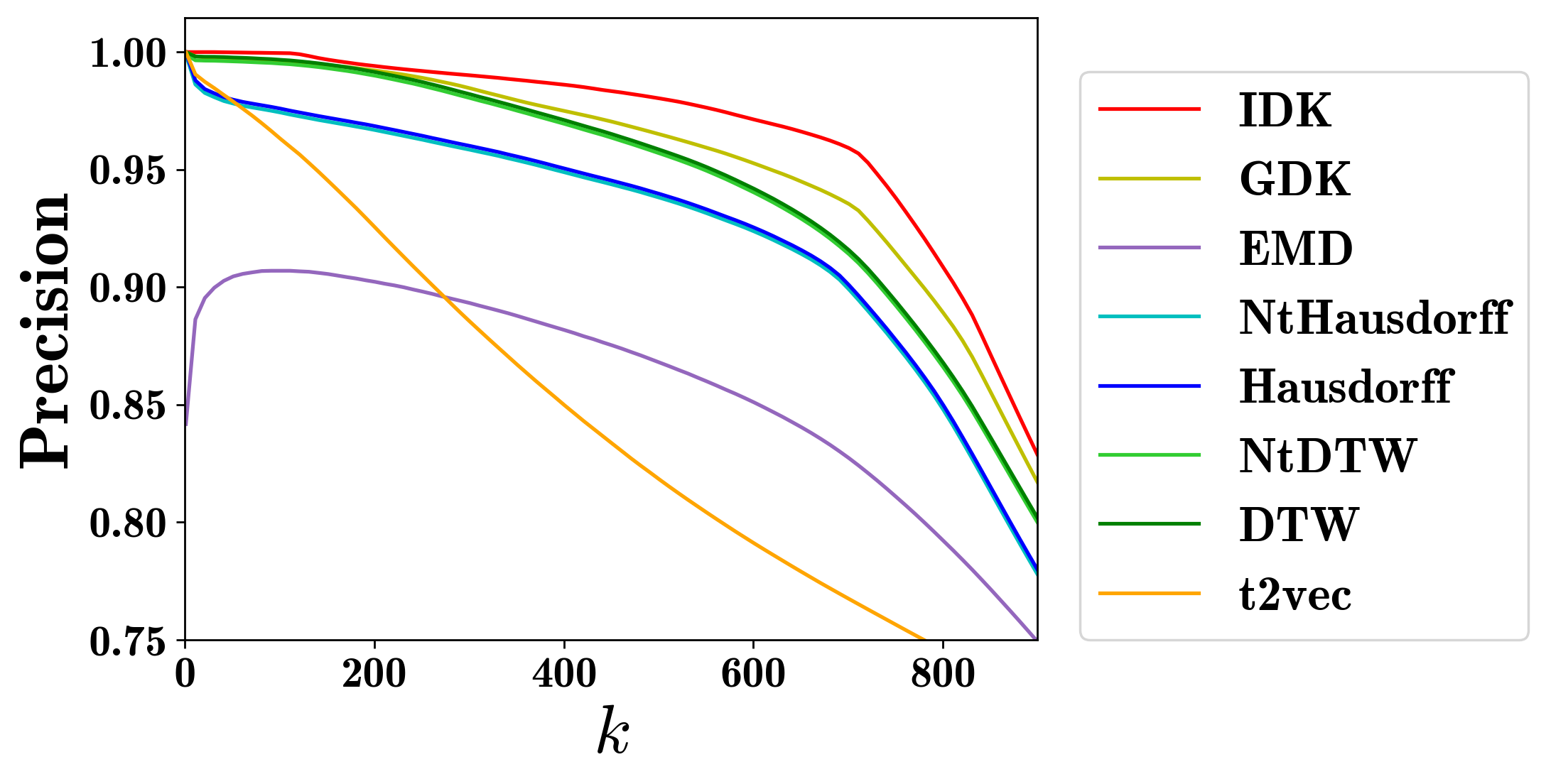}
    \caption{Precision@$k$ with different distance measures on the Geolife dataset. }
    \label{fig: TopK_geolife}
\end{figure}

\paragraph{Retrieval Evaluation}
Here we assess the goodness of a trajectory distance measure in a trajectory retrieval task.
Every trajectory in a dataset is used as a query, and we report the average precision of $k$ nearest trajectories (precision@$k$) as the final result. The retrieval results in Figure~\ref{fig: TopK_geolife} show that IDK has the best retrieval outcomes over all other measures for every $k$ value. GDK, DTW and NtDTW are the next best measures. NtHausdorff and NtDTW are deep learning methods that approximate Hausdorff distance and DTW distance, respectively.  These two methods provide good approximations of the two distances with almost the same precision@$k$ as the original distance calculation method (their lines almost overlap in Figure~\ref{fig: TopK_geolife}), but they could not be better than the original distances. The precision of EMD is the lowest of all measures.
This is mainly due to the fact that EMD uses a grid to partition the trajectories, which leads to a reduction in the accuracy of feature extraction, where trajectories in the same grid cell have zero distances.  

Note that all measures have a significant drop as $k$ increases beyond 720. This is because the average size of a cluster is 766. When the size of $k$ exceeds the size of a cluster, at least one retrieved trajectory is from a different cluster.

IDK has been shown to be a better measure than GDK in point and group anomaly detection \cite{KaiMingTing2020IsolationDK} and time series anomaly detection \cite{IDK-timeseries-VLDB2022}. Our result in trajectory retrieval is consistent with these results.

\section{Isolation distributional kernel clustering}
\label{TIDKC}


\begin{figure*}[!tb]
  \centering
  \includegraphics[width=0.95\linewidth]{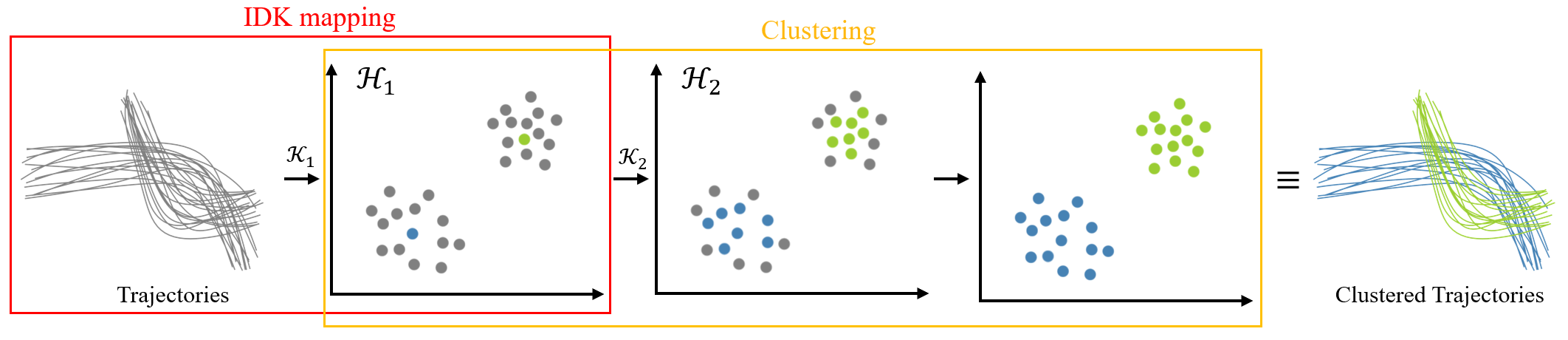}
  \caption{The top-level view of TIDKC trajectory clustering based on IDK.}
  \label{fig:procedure}
\end{figure*}

In this section, we propose a trajectory clustering using IDK, called TIDKC (Trajectory Isolation Distributional Kernel Clustering). The overview of the clustering procedure is presented in Figure \ref{fig:procedure}. The detailed procedure is provided in Algorithm \ref{alg1}. 
TIDKC uses two levels of IDK. The level-1 IDK $\mathcal{K}_1$ is derived from $S=\cup_i T_i$, and its feature map $\widehat{\Phi}_1(\cdot|S)$ is used to map each trajectory $T_i$ to a level-1 point $g_i = \widehat{\Phi}_1(\mathcal{P}_{T_i} |S)=\frac{1}{|T_i|} \sum_{x \in T_i} \phi(x|S)$ in RKHS $\mathcal{H}_1$ to yield $G=\{g_1, \dots, g_n \}$ (line 1 in Algorithm~\ref{alg1}), where $\phi(\cdot|S)$ is the feature map of Isolation Kernel  \cite{KaiMingTing2018IsolationKA}. 

The clustering, in the following step, is performed on set $G$ by growing each cluster (of level-1 points in $G$) via level-2 IDK $\mathcal{K}_2$ in RKHS $\mathcal{H}_2$. The idea is to identify $k$ seeds
to initialize the $k$ clusters $C_j$, represented as distribution $\mathsf{P}_{C_j}$, which are then grown up to a similarity threshold $\tau$. As the clusters have grown, they are used as the new basis to grow further to a lower threshold. This repeats until all points in $G$ are assigned to one of the $k$ clusters. Lines 5 to 9 in Algorithm~\ref{alg1} show the procedure of the clustering.

\begin{algorithm}[!t]  
\footnotesize
	\caption{TIDKC: Trajectory clustering using IDK} 
	\label{alg1} 
\textbf{Input:} $D= \{T_1, \dots, T_n\}$: trajectory dataset,  
		$k$: number of clusters,  $\varrho \in (0,1)$: growth rate of clusters \\
 \textbf{Output:} $C= \{E_1, \dots, E_k\}$: $k$ Clusters.
	\begin{algorithmic}[1]
		\STATE Map each trajectory $T_i\in D$ to a point $g_i=\widehat{\Phi}_1(\mathcal{P}_{T_i})$ in RKHS $\mathcal{H}_1$ using feature map $\widehat{\Phi}_1$ of $\mathcal{K}_1$ to yield $G=\{g_1, \dots, g_n \}$
      \STATE Select $k$ cluster seeds $c_j \in G$ based on local-contrast estimation; and initialize clusters $C_j = \{c_j\}, j=1,\dots,k$
      \STATE Initialize $N \leftarrow G \setminus \cup_{j} C_j$
\STATE Initialize $\tau \leftarrow \displaystyle \max_{{g} \in N, \ell \in [1,k]} \mathcal{K}_2(\delta(g),\mathsf{P}_{C_\ell})$
	 \REPEAT
	 \STATE $\tau \leftarrow  \varrho \times \tau $
	 \STATE Expand cluster  $C_j$ to include unassigned point $g\in N$ for\\ $j=\argmax_{\ell \in [1,k]} \mathcal{K}_2(\delta(g),\mathsf{P}_{C_\ell})$ and $\mathcal{K}_2(\delta(g),\mathsf{P}_{C_j}) > \tau$
  \STATE	$N \leftarrow G \setminus \cup_{j} C_j$
 	\UNTIL $|N|=0$  or $\tau<0.00001$
    \STATE Assign each unassigned point $g$ to nearest cluster $C$ via $\mathcal{K}_2(\delta(g),\mathsf{P}_{C})$
    \STATE Cluster $E_j \subset D$ corresponds to $C_j \subset G, j=1,\dots,k$
	\end{algorithmic} 
\end{algorithm}

\begin{table}[!t]
 \centering
 \caption{Time complexities of five main processes in TIDKC. $t$ and $\psi$ are parameters of Isolation Kernel \cite{KaiMingTing2018IsolationKA}. $d$, $k$ and $b$ are numbers of dimensions,  clusters and iterations, respectively.} 
\label{tab:time_complexity}
  \begin{tabular}{llr}
  \toprule
1. & Build Isolation Kernel (IK)   & $\mathcal{O}(dt\psi)$\\
2. & Getting the feature map   & $\mathcal{O}(n dt\psi)$ \\ 
3. & Cluster seeds selection & $\mathcal{O}(s^2t\psi)$ \\
4. &  Growing $k$ clusters  &  $\mathcal{O}(bknt\psi)$  \\
5.  &  Final clusters assignment  &  $\mathcal{O}(knt\psi)$  \\
\midrule
\multicolumn{2}{l}{Total time cost for TIDKC}  & $\mathcal{O}((s^2+n(bk+d))t\psi)$\\ 
  \bottomrule
 \end{tabular}
\end{table}

The level-2 IDK $\mathcal{K}_2$ is derived from $G$, and it is used to perform the clustering on $G$. The idea is to identify $k$ seeds\footnote{We utilise a local contrast method \cite{chen2018local} to identify the $k$ local peaks as cluster seeds.} to initialize the $k$ clusters $C_j$, represented as distribution $\mathsf{P}_{C_j}$, which are then grown up to a similarity threshold $\tau$. As the clusters have grown, they are used as the new basis to grow further to a lower threshold. This repeats until all points in $G$ are assigned to one of the $k$ clusters.


The final clustering outcome is clusters $E_j$ of trajectories which correspond to clusters $C_j$ of the level-1 IDK-mapped points in RKHS $\mathcal{H}_1$ for $j=1,\dots,k$.

\textbf{Clustering objection function}:
Given a dataset of level-1 mapped points $G=\{g_1,\dots,g_n\}$,
TIDKC aims to find clusters $C_j, j=1,\dots,k$ satisfying the following objective function using level-2 $\mathcal{K}_2$:
\begin{equation}
\argmax_{C_1,\dots,C_k} \sum^k_{j=1} \sum_{g \in C_j} \mathcal{K}_2(\delta(g),\mathsf{P}_{C_j}|G),
\label{eqn_maximizing-function}
\end{equation}
\noindent 
where $\delta(\cdot)$ is a Dirac measure and $\mathsf{P}_C$ is the distribution that generates the set $C$ of points in RKHS $\mathcal{H}_1$.

Note that TIDKC is developed based on IDKC~\cite{IDKC} for trajectory clustering. IDKC is the first clustering algorithm that employs an adaptive distributional kernel without any optimisation. IDKC is able to identify complex clusters on massive datasets, but it works on multi-dimensional data points only. The key additional step is line 1 in Algorithm~\ref{alg1}.

Table \ref{tab:time_complexity} presents the time complexities of the key procedures of TIDKC.  Building IK and mapping $n$ points using IK's feature map cost $\mathcal{O}(n d \psi t)$, where $\psi \ll n$. The cluster seeds selection can be based on a  subset of fixed size $s \ll n$ for a large dataset. 
The cluster growing process is based on the distributional kernel with $\mathcal{O}(n)$. The number of iterations for lines 5 to 9 in Algorithm \ref{alg1} is bounded by the growth rate ($\varrho$ in line 6). Thus, the total time complexity of TIDKC is linear to the dataset size $n$ since all other parameters are constant.

\section{Gaussian distributional kernel clustering}
\label{TGDKC}
Note that it is possible to produce different variants of TIDKC by simply replacing IDK in $\mathcal{K}_1$ or/and $\mathcal{K}_2$ with a Gaussian distributional kernel (GDK). 

To produce TGDKC (Trajectory Gaussian Distributional Kernel Clustering), IDK is replaced with GDK as $\mathcal{K}_2$ in Algorithm \ref{alg1}, where the cluster growing step is guided by GDK instead.

Both TGDKC and TIDKC can use either IDK or GDK as $\mathcal{K}_1$ in Algorithm \ref{alg1} for trajectory representation and similarity measurement.

All these variants are shown in Table \ref{tab:variants}. Note that all these variants use the same Algorithm \ref{alg1}, except for different combinations of kernels.
In order to examine the effectiveness of IDK in $\mathcal{K}_1$ or/and $\mathcal{K}_2$, we apply both TIDKC and TGDKC and their variants in the evaluation, reported in Section~\ref{Eva}.

\begin{table}[t]
  \centering
  \caption{Variants of TGDKC and TIDKC.}
    \begin{tabular}{lcc}
    \toprule
          & TGDKC  & TIDKC \\
    \midrule
    $\mathcal{K}_1$ (Trajectory representation)   & IDK/GDK & IDK/GDK \\
    $\mathcal{K}_2$ (Growing Clusters)    & GDK   & IDK  \\
    \bottomrule
    \end{tabular}%
  \label{tab:variants}%
\end{table}%

When computing TGDKC, we apply a Nyström method  to speed up its similarity computation\footnote{The Nyström method  \cite{KaiXuanChen2018RiemannianKB} constructs a low rank feature map to approximate the original kernel matrix by sampling, and reduces the computational cost of the kernel matrix in the calculation.}. The time complexity of TGDKC is $\mathcal{O}((s^2+ n(bk+m)+l^2)l)$, where $m$ and $l$ are the sample size and the dimensionality of the approximate feature map used in the Nyström method, respectively. Like TIDKC, TGDKC is also a linear time algorithm because all parameters, except $n$, are constant.

\begin{table}[b]
    \centering
    \footnotesize
    \caption{Real-world datasets. Ave $|T|$ indicates the average length of trajectories.}
    \label{datainfo}
    \begin{tabular}{@{}crrrrr@{}}
    \toprule
    Dataset & \#Points & \#Traj & min--max $|T|$ & Avg $|T|$ & \#Clusters \\
    \midrule
        VRU\_pedes\_3   & 202,272 & 610 & 196--620 & 332 & 3 \\
        VRU\_pedes\_4   & 238,443 & 710 & 196--620 & 336 & 4 \\
        VRU\_cyclists   & 76,051 & 265 & 52--1,257 & 287 & 3 \\
        TRAFFIC & 15,000 & 300 & 50--50 & 50 & 11 \\
        CROSS & 24,420 & 1,900 & 5--23 & 13 & 19 \\
        CASIA & 147,883 & 1,500 & 16--612 & 96 & 15 \\
        Geolife & 620,477 & 9,192 & 24--100 & 68 & 12\\
    \bottomrule
    \end{tabular}
\end{table}

\section{Empirical Evaluation}
\label{Eva}

\subsection{Experimental design and settings}

In this section, we aim to answer the following questions:
\begin{enumerate}
    \item Is distributional kernel an effective means to represent trajectories and compute the similarity between trajectories in clustering tasks?
    \item Does TIDKC have any advantage over other clustering methods on trajectory clustering, in terms of clustering performance and time complexity?
\end{enumerate}

\begin{table*}[t]
    \centering
    \small
    \caption{Clustering outcomes in terms of NMI (Normalized Mutual Information \cite{cover1999elements}).} 
    \label{tab_NMI}
    \begin{tabular}{@{}c|ccccc|ccccc|cc|cc|c@{}}
    \hline
    \multirow{2}{*}{}&\multicolumn{5}{c|}{K-Medoids clustering} & \multicolumn{5}{c|}{spectral clustering} & \multicolumn{2}{c|}{TIDKC} & \multicolumn{2}{c|}{TGDKC} &  \multirow{2}{*}{E$^2$DTC} \\
    \cline{2-15}
     & IDK & GDK & HD & DTW & t2vec & IDK & GDK & HD & DTW & t2vec & IDK & GDK & IDK & GDK \\
        \hline
       VRU\_pedes\_3 & .74 & .72 & .64 & .52 & .48 & .85 & .83 & .85 & .55 & .58 & .94 & \textbf{.97} & .94 & .80 & .69\\ 
        VRU\_pedes\_4 & .55 & .51 & .40 & .50 & .41 & .57 & .60 & .67 & .62 & .51 & \textbf{.84} & .80 & .70 & .65  &  .51\\ 
        VRU\_cyclists & .75 & .46 & .64 & .75 & .56 & .74 & .58 & \textbf{.94} & .74 & .64 & .83 & .67 & .61 & .66 & .86\\ 
        TRAFFIC & .91 & .91 & .95 & .96 & .72 & .96 & .97 & .69 & .77 & .61 & \textbf{1.00} & .98 & \textbf{1.00} & .96 & .95\\ 
        CROSS & .98 & .98 & .97 & .97 & .97 & .97 & .95 & .85 & .61 & .73 & \textbf{.99} & .98 & .95 & .95 & .91\\ 
        CASIA & .91 & .92 & .90 & .92 & .80 & \textbf{.94} & .92 & .86 & .85 & .83 & .91 & .90 & .89 & .89 & .76\\ 
        Geolife & .92 & .89 & .75	& .82 & .86 & .95 & .89 & .93 & .97 & .90 & \textbf{.99} & .96 & .96 & .97 & .79\\\hline
        Average NMI & .81 & .76 & .73 & .77 & .70 & .83 & .80 & .80 & .73 & .71 & \textbf{.91} & .86 & .85 & .83 & .80 \\
    \hline
    \end{tabular}
\end{table*}

\begin{table*}[t]
    \centering
    \small
    \caption{Clustering outcomes in terms of ARI (Adjusted Rand Index \cite{steinley2004properties}).}
    \label{ARI}
    \begin{tabular}{@{}c|ccccc|ccccc|cc|cc|c@{}}
    \hline
    \multirow{2}{*}{}&\multicolumn{5}{c|}{K-Medoids Clustering} & \multicolumn{5}{c|}{spectral Clustering} & \multicolumn{2}{c|}{TIDKC} & \multicolumn{2}{c|}{TGDKC} &  \multirow{2}{*}{E$^2$DTC}\\
    \cline{2-15}
     & IDK & GDK & HD & DTW & t2vec & IDK & GDK & HD & DTW & t2vec & IDK & GDK & IDK & GDK \\
    \hline
        VRU\_pedes\_3 & .74 & .70 & .60 & .47 & .46 & .85 & .78 & .82 & .52 & .59 & .97 & \textbf{.98} & .88 & .71  & .67\\
        VRU\_pedes\_4 & .49 & .47 & .34 & .55 & .31 & .54 & .51 & .63 & .60 & .43 & \textbf{.85} & .74 & .56 & .45 & .50\\ 
        VRU\_cyclists & .78 & .39 & .55 & .68 & .57 & .75 & .61 & \textbf{.95} & .63 & .54 & .79 & .58 & .53 & .51 & .90\\ 
        TRAFFIC & .87 & .86 & .95 & .94 & .47 & .95 & .96 & .35 & .46 & .32 & \textbf{1.00} & .97 &\textbf{1.00} & .94 & .91\\ 
        CROSS & .97 & .97 & .97 & .96 & .97 & .98 & .92 & .56 & .32 & .29 & \textbf{.98} & \textbf{.98} & .92 & .91 & .78\\ 
        CASIA & .83 & .82 & .80 & .85 & .64 & .90 & .89 & .70 & .69 & .70 & \textbf{.93} & .90 & .83 & .79 & .54\\ 
        Geolife & .73 & .68 & .43 & .60 & .77 & .88 & .60 & .82 & .92 & .83 & \textbf{.99}  & .89  & .90  & .91  & .52\\\hline
        Average ARI & .74 & .68 & .63 & .69 & .62 & .80 & .73 & .66 & .58 & .57 & \textbf{.90} & .84 & .78 & .73 & .71\\ 
     \hline
    \end{tabular}
\end{table*}

To answer the first question, we compare IDK and GDK with traditional trajectory distance measures\footnote{Fréchet distance and Earth mover's distance are omitted because they took extremely long time to run, especially when a trajectory has a large number of points.},
including Hausdorff distance \cite{AbdelAzizTaha2015AnEA} and DTW distance \cite{RakeshAgrawal1993EfficientSS}, and the deep learning method t2vec \cite{XiuchengLi2018DeepRL}. They are evaluated in the same clustering algorithm.
To answer the second question, we compare TIDKC and TGDKC with three existing clustering methods, i.e., K-Medoids clustering  \cite{sklearn_api}, Spectral clustering  \cite{sklearn_api}, and an end-to-end deep learning clustering method E$^2$DTC  \cite{ZiquanFang2021E2D}. We search their parameters following the original papers and report their best performances.

Because K-Medoids and spectral clustering accept a distance matrix as input, all five measures are used to answer the first question (i.e., IDK, GDK, Hausdorff, DTW and t2vec). 
As both TIDKC and TGDKC require  a multi-dimensional representation of trajectories and do not accept a distance matrix as input, only IDK and GDK 
are used for trajectory similarity measurements. 

Table \ref{datainfo} presents the datasets used in the experiments. Most of the datasets are trajectories of vehicles or pedestrians recorded by GPS or surveillance cameras. They vary in size, lengths of trajectories and number of clusters. 


\subsection{Clustering performance}

The clustering performances of  five clustering methods using different distance measures are shown in Table \ref{tab_NMI} (in terms of NMI) and Table \ref{ARI} (in terms of ARI). 

We have the following observations:

\begin{enumerate}
\item 
Comparing among the five distance measures, IDK has the best average NMI and ARI using either K-Medoids or spectral clustering. 
Overall, the closest contender to IDK is GDK for these two clustering methods.

\item Deep representing method t2vec performed the worst on most datasets  using either K-Medoids or Spectral clustering. It has mixed results between these two clustering methods on different datasets. 

\item
TIDKC with either IDK or GDK outperformed all other clustering methods on almost all datasets.  It is interesting to note that TGDKC, spectral clustering and K-Medoids performed the best with IDK. 
\item The end-to-end deep learning clustering method, E$^2$DTC, has some improvement over Spectral clustering and K-Medoids, only if the latter two use DTW and t2vec; but E$^2$DTC always performed worse than TIDKC with one exception.
\end{enumerate}

In a nutshell, TIDKC, in combination with IDK to represent and compute the similarities of trajectories, offers the best method for trajectory clustering.


\subsection{Scaleup tests}



\begin{figure*}[!tb]
	\centering
	\begin{subfigure}{0.42\textwidth}
		\centering
\includegraphics[width=0.9\linewidth]{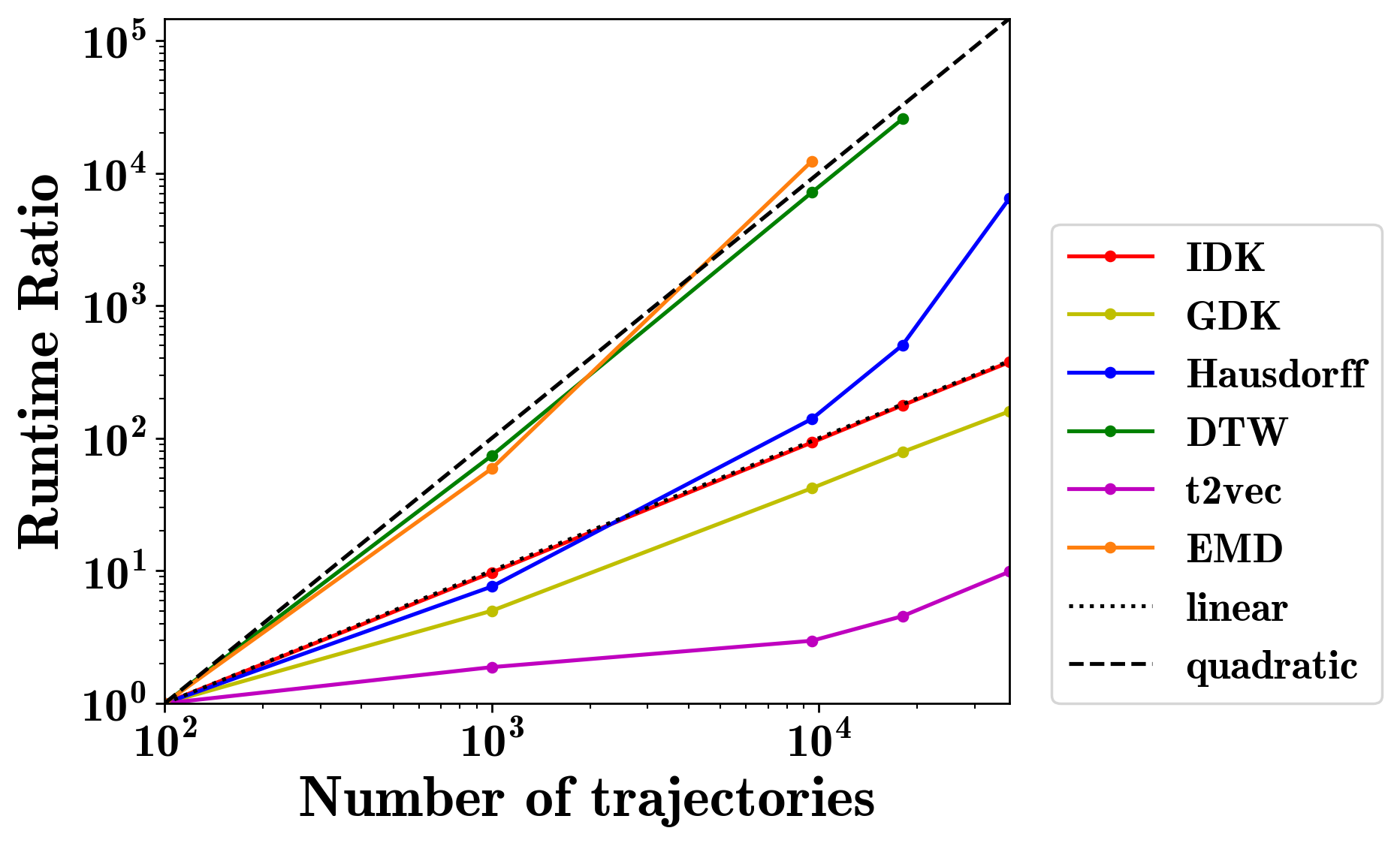}
\caption{Scaleup test for different distance/kernel measures.}
		\label{fig:scaleupDistance}
	\end{subfigure}  
	\begin{subfigure}{0.42\textwidth}
		\centering
\includegraphics[width=0.9\linewidth]{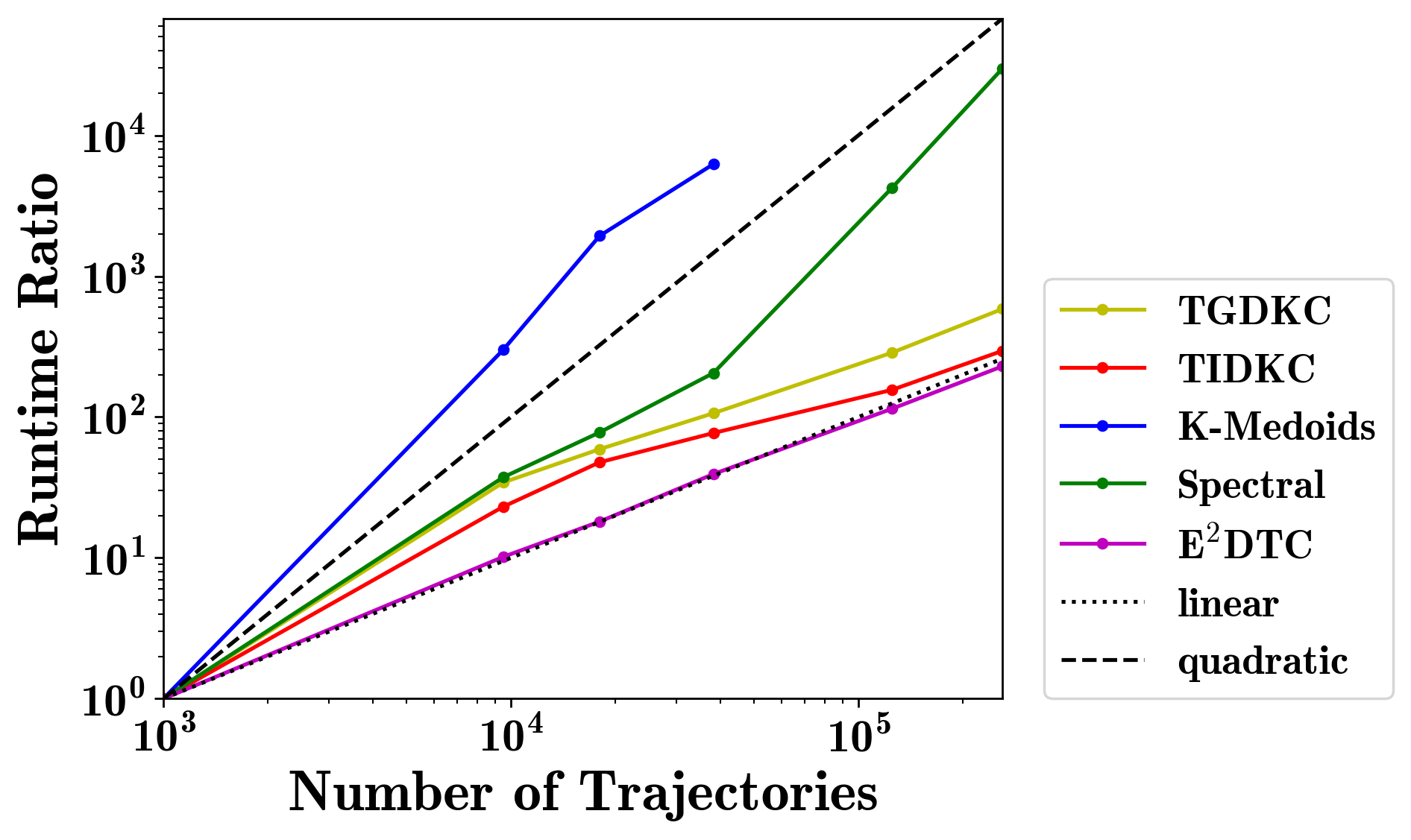}
\caption{Scaleup test for different clustering methods.}
		\label{fig:scaleupCluster}
	\end{subfigure} 
	\caption{Scaleup tests.
  All measures and clustering methods  ran on CPU, except t2vec and E$^2$DTC which ran on GPU. All clustering methods use IDK as the trajectory representation and distance measurement, except E$^2$DTC.}
	\label{fig:sca}
\end{figure*}

\begin{figure*}[!tb]
\centering
\begin{subfigure}{0.3\textwidth}
		\centering
		\includegraphics[width=0.85\textwidth]{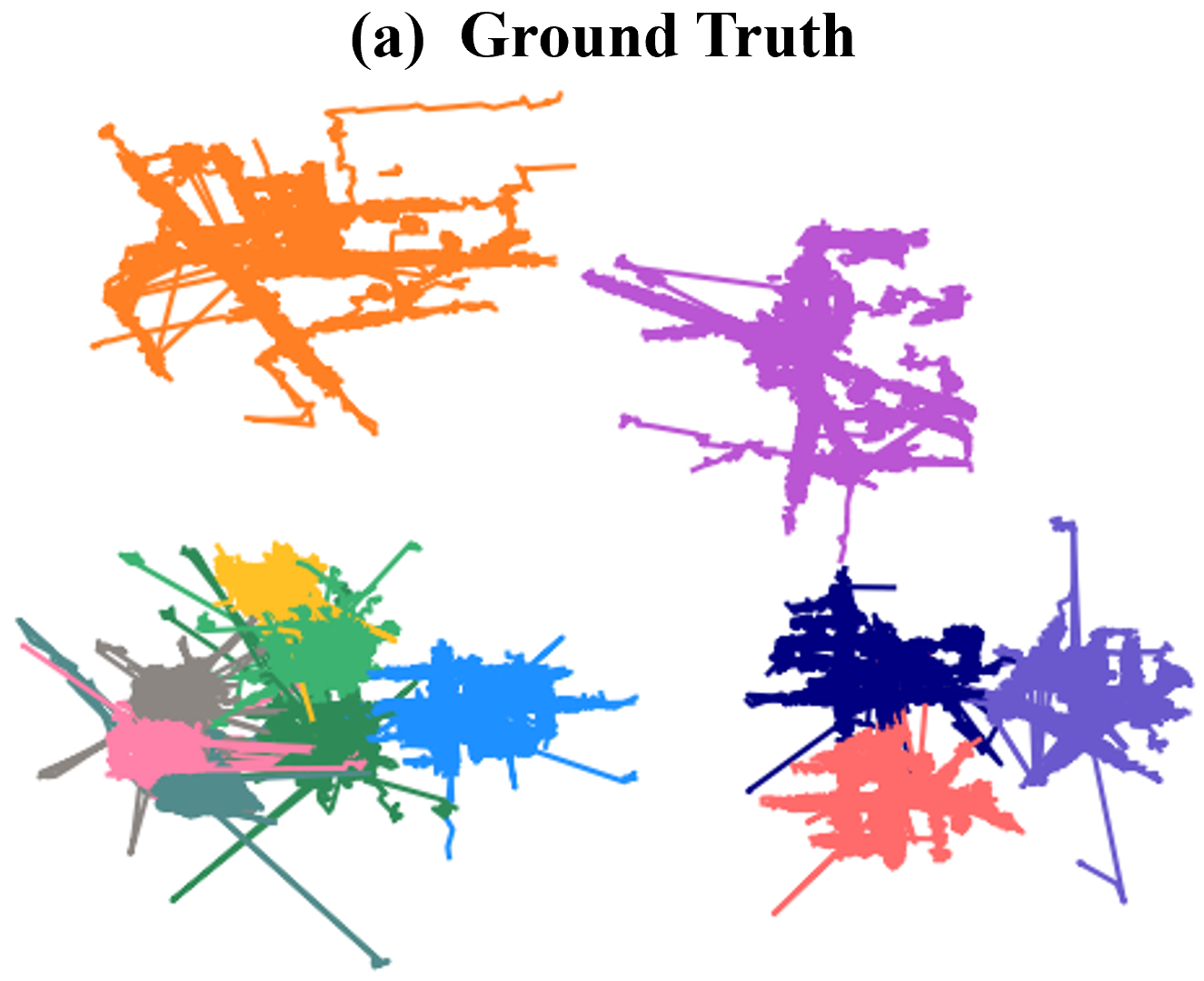}
	\end{subfigure}
	\begin{subfigure}{0.3\textwidth}
		\centering
		\includegraphics[width=0.85\textwidth]{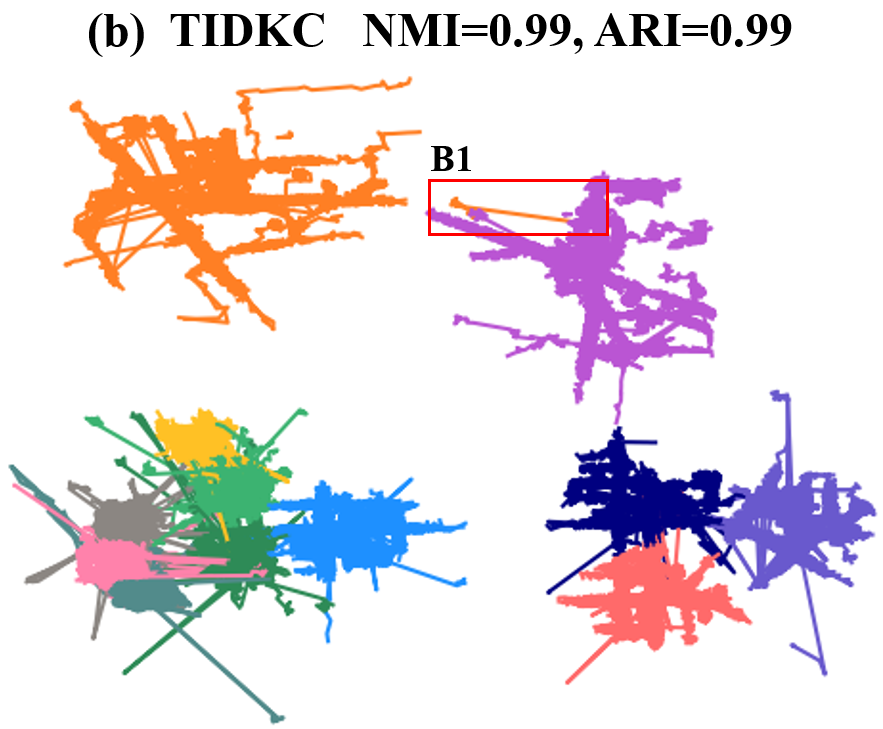}
	\end{subfigure}\\
	\begin{subfigure}{0.3\textwidth}
		\centering
		\includegraphics[width=0.85\textwidth]{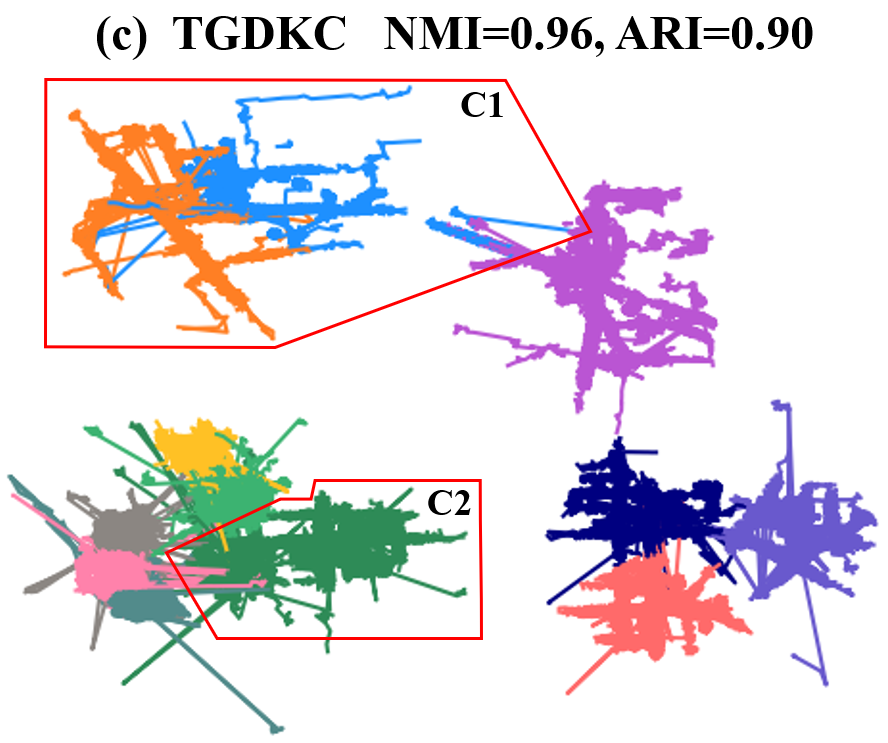}
	\end{subfigure}
	\begin{subfigure}{0.3\textwidth}
		\centering
		\includegraphics[width=0.85\textwidth]{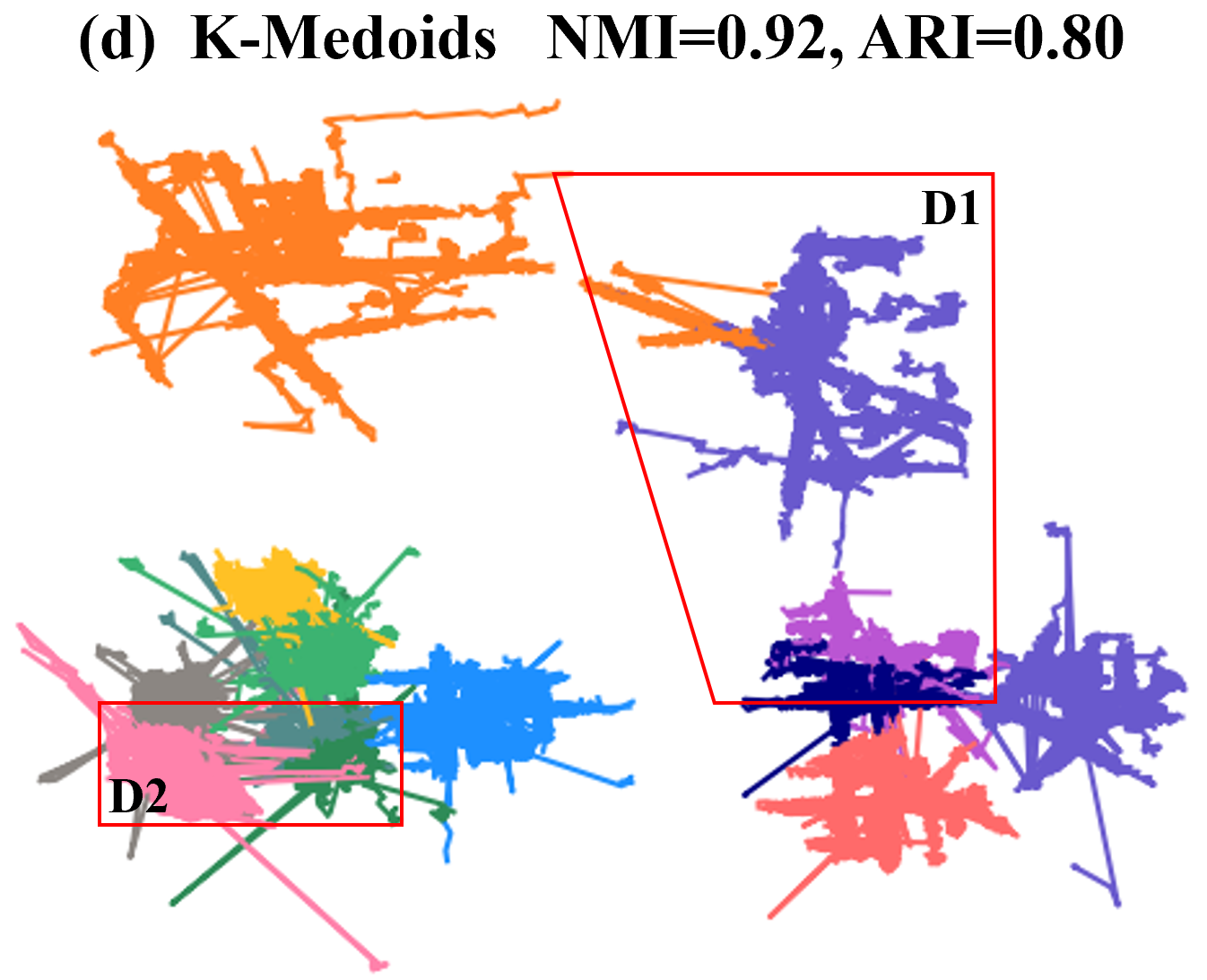}
	\end{subfigure} 
	\begin{subfigure}{0.3\textwidth}
		\centering
		\includegraphics[width=0.85\textwidth]{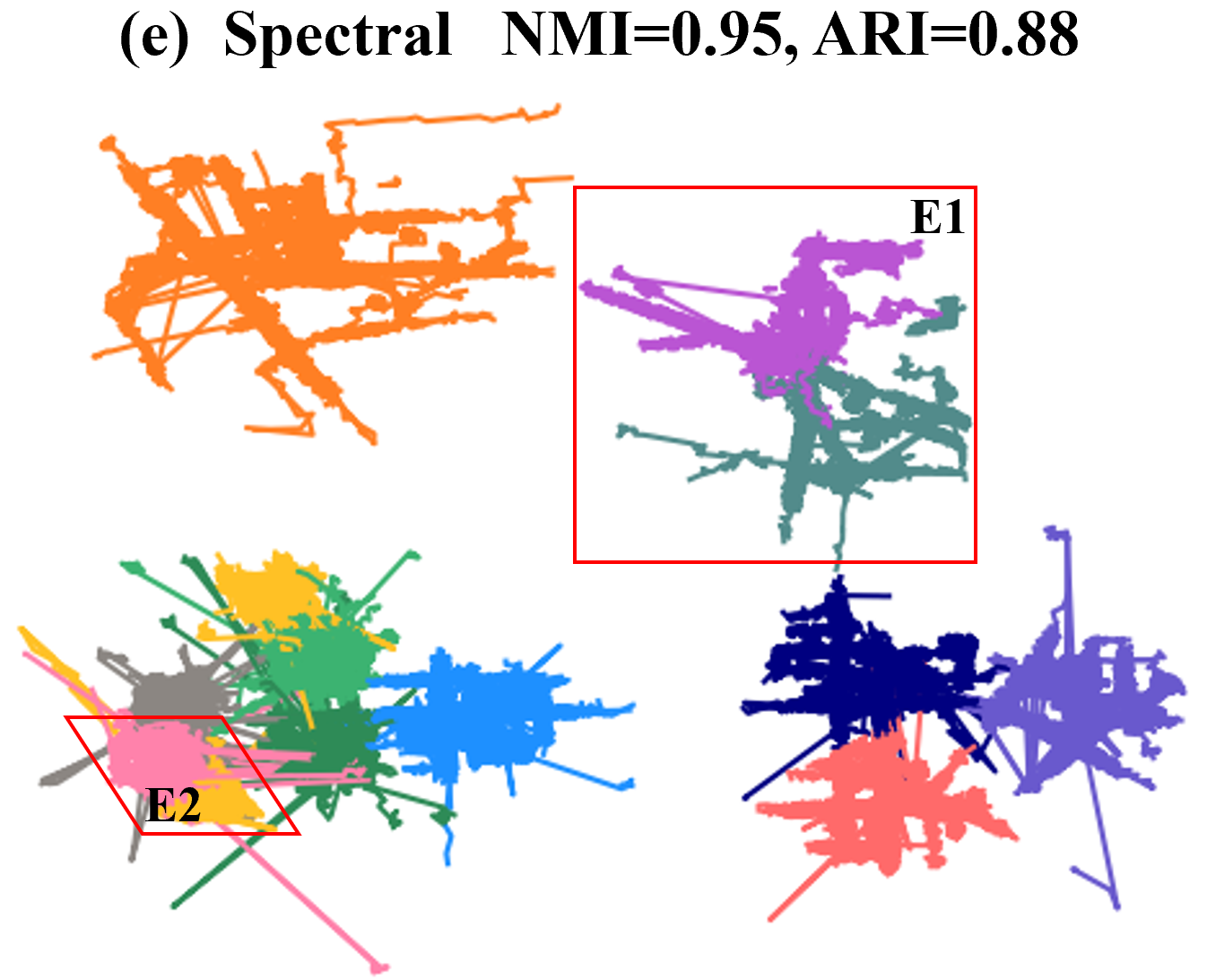}
	\end{subfigure}
	\caption{Clustering outcomes of clustering methods using the same representation of IDK on the Geolife dataset.  Areas with clustering errors are indicated by the red boxes.}
	\label{fig:geolifeResults}
\end{figure*}

We conduct  scaleup tests on both distance measures and clustering methods\footnote{Multi-thread processing is applied in Hausdorff and DTW distances, but not others. All ran in CPU, except that t2vec and E$^2$DTC ran in GPU.}. The dataset of scaleup tests is constructed based on the Geolife dataset. \autoref{fig:scaleupDistance} shows the scaleup test result of different distance measures. 
Because of the use of GPU, t2vec has the lowest runtime ratio. Apart from t2vec, GDK with the Nyström method, has the best scaleup ratio, followed by IDK with linear runtime. Hausdorff and DTW distances have at least quadratic runtime, making them the slowest measures. 
The deep learning approximations for Hausdorff and DTW distance (NtHausdorff and NtDTW) have linear time. The lines are omitted for brevity.


\autoref{fig:scaleupCluster} shows the scaleup test result of different clustering methods. K-Medoids is the slowest with quadratic runtime. Spectral clustering also has super-linear time complexity, especially when the dataset is large. TIDKC and TGDKC have similar linear runtime ratios with TIDKC being slightly faster. TIDKC and TGDKC have slightly different runtimes though they use the same algorithm. This is because the runtime of TGDKC or TIDKC depends on the number of iterations required to complete the task, mainly due to the kernel (GDK or IDK) employed. 
E$^2$DTC has the lowest runtime ratio with the use of GPU.

\subsection{Clustering Visualisation}
We provide exactly the same IDK similarities on the Geolife dataset to TIDKC, TGDKC, K-Medoids and spectral clustering to investigate their clustering capabilities. This is shown in \autoref{fig:geolifeResults}. Any clustering mistakes are a direct result of a clustering method since the trajectory similarities on this dataset, provided by IDK, enable an ideal clustering method to produce a perfect clustering outcome. The clustering outcomes of TIDKC, TGDKC, K-Medoids clustering and spectral clustering are shown in subfigures 5(b), (c), (d) \& (e), respectively.

Our observations are given as follows:
\begin{enumerate}
    \item TIDKC performs very well on the dataset, clustering almost every trajectory correctly, except for the trajectory in region B1.
    \item TGDKC is less accurate in dealing with clusters of different densities, compared to TIDKC. For clusters in sparse regions like region C1 in subfigure (c), TGDKC tends to divide the points in the same cluster into different subclusters. For dense clusters, TGDKC might merge two different clusters into one, as shown in region C2 in subfigure (c).
    \item K-Medoids clustering relies heavily on the initialization of centroids and can only handle the clustering of globular shapes (in Hilbert space induced by IDK). As a result, it splits one cluster into two subclusters for two clusters in region D1; and merges two clusters into one in region D2, as shown in subfigure (d). 
    \item Spectral clustering has clustering errors in region E1 and region E2, shown in subfigure (e). These errors can be attributed to the fundamental limitation of SC identified previously on clusters of varied densities (see \cite{BoazNadler2006FundamentalLO} for details).
\end{enumerate}

\subsection{IDK representation with time order information}

The IDK similarity calculation in Equation \ref{IDKmearue} is based on the distribution of spatial points only, without considering the temporal information. In the previous experiments, since there is no time/direction information in the datasets, we use spatial information only.

However, some trajectories could be similar, except travelling in opposite directions. Since DTW and t2vec handle temporal information, they are able to differentiate trajectories with different directions. While basic IDK considers spatial distributions only, in applications where separating these trajectories into different clusters is important, it can also be used by simply adding one dimension to represent the temporal order of the points of each trajectory. 

\begin{figure}[!tb]
    \centering
    \includegraphics[width=\linewidth]{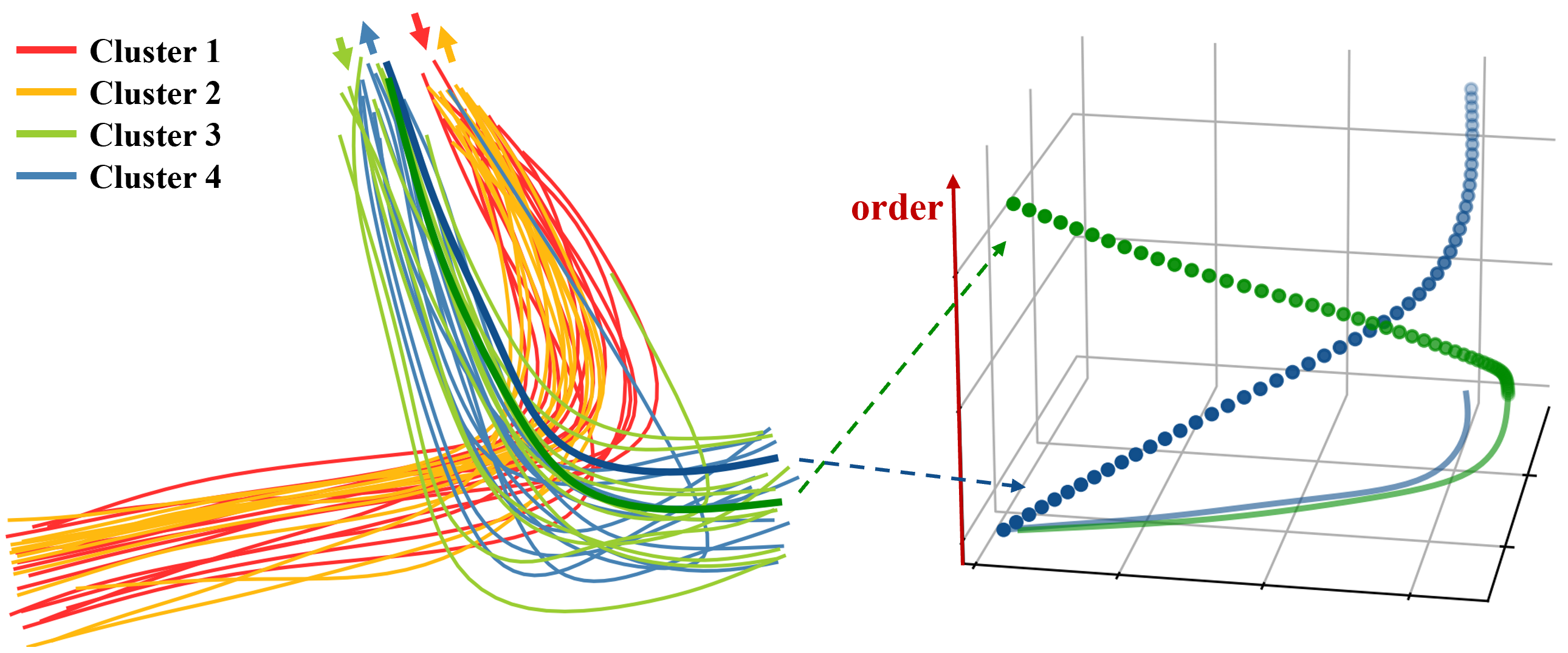}
    \caption{The left subfigure shows the simulated direction-sensitive 4 clusters of the TRAFFIC dataset. Clusters 1 \& 2 have opposite travelling directions; so as Clusters 3 \& 4 in the original 2-dimensional space. The right subfigure shows two example trajectories of similar shapes but opposite directions in the 3-dimensional space after adding an order dimension.}
    \label{fig:timetraj}
\end{figure}

Here we simulate such a scenario by modifying two clusters of trajectories from the TRAFFIC dataset as an example. We randomly picked half of the trajectories in a cluster and then assigned them the direction from top to bottom; while the remaining trajectories are assigned an opposite direction. As a result, there are four clusters when considering both the directions and trajectory patterns. To deal with such problems, we add a dimension to each trajectory to represent the order of the points, thus transforming the trajectory from a two-dimensional distribution to a three-dimensional one. As shown in \autoref{fig:timetraj}, for two trajectories that are very close in shape but opposite in direction, their distributions in 3D space are very different after the addition of the order dimension, and thus trajectories of similar shapes and opposite directions can be easily clustered into two clusters by TIDKC. It is worth noting that the order dimension we have added is not the original timestamp of the trajectory. Therefore the timestamps of the original trajectories  have no impact on the clustering results.

With the above-stated order dimension added, TIDKC is able to distinguish the four clusters without errors, yielding NMI=1. Note that without the order dimension, TIDKC  has NMI=0.55.


\subsection{Influence of sampling rate on trajectory clustering}

Trajectories may have different sampling rates due to differences in sampling equipment. In this section, we examine the effect of different sampling rates on trajectory clustering. In the first experiment, we downsample all of the trajectories in the dataset and apply trajectory clustering methods. In the second experiment, we examine the effect of varying sampling rates in the same dataset. For each cluster in the dataset, we randomly select half of the trajectories and downsample them. Therefore, one cluster would contain two sets of trajectories with different sampling rates. We gradually reduce the sampling rate and examine the clustering results of different trajectory clustering methods. We conduct experiments on Geolife dataset, using IDK, GDK and Hausdorff distance to measure the similarity of trajectories and then applying spectral clustering. Results are shown in \autoref{tab: sample rate}. We can see that no matter which similarity measure is used, the NMI scores do not degrade significantly as the sampling rate decrease, which suggests that changes in sampling rate do not have a significant effect on the clustering results on the Geolife dataset.

\begin{table}[!tb]
\centering
\small
\caption{NMI score of spectral clustering with IDK, GDK and Hausdorff distance on the Geolife dataset with different sampling rates. `All' means the entire dataset is downsampled, while `Half' means only half of the trajectories in the dataset are downsampled.}
\label{tab: sample rate}
\begin{tabular}{@{}c|ccccc@{}}
\hline
Sampling rate    & 1   & 0.9 & 0.7 & 0.5 & 0.3 \\
\hline
All - IDK        & .95 & .95 & .93 & .93 & .92 \\
All - GDK        & .89 & .88 & .87 & .85 & .82 \\
All - Hausdorff  & .93 & .93 & .92 & .91 & .89 \\
\hline
Half - IDK       & .95 & .95 & .93 & .92 & .92 \\
Half - GDK       & .89 & .87 & .85 & .85 & .81 \\
Half - Hausdorff & .93 & .92 & .91 & .88 & .88 \\ \hline
\end{tabular}
\end{table}

\section{Conclusion}
\label{sec-conclusions}
In this paper, we found that IDK is a powerful tool for trajectory representation and similarity measurement as well as trajectory clustering. 
The proposed TIDKC  makes use of distributions at two levels of IDK. The first level IDK is used to represent each trajectory as a distribution. The second level IDK is used to represent the distribution of a cluster of trajectories, enabling clusters with irregular shapes and varied densities to be discovered. 

Our extensive evaluation confirms that IDK is more effective in capturing complex structures in trajectories than traditional and deep learning-based distance measures, such as DTW, Hausdorff distance and t2vec. Furthermore, the proposed TIDKC has superior clustering performance in terms of NMI than existing trajectory clustering algorithms, including K-Medoids clustering, spectral clustering, TGDKC and E$^2$DTC. With linear runtime, IDK is faster than point-based trajectory distance measures like DTW, Hausdorff distance and distribution-based method EMD. TIDKC runs orders of magnitude faster than other methods except deep learning-based that was accelerated by GPU.

\bibliographystyle{splncs04}
\bibliography{sample}

\begin{thebibliography}{10}
\providecommand{\url}[1]{\texttt{#1}}
\providecommand{\urlprefix}{URL }
\providecommand{\doi}[1]{https://doi.org/#1}

\bibitem{RakeshAgrawal1993EfficientSS}
Agrawal, R., Faloutsos, C., Swami, A.: Efficient similarity search in sequence databases. In: Foundations of Data Organization and Algorithms, pp. 69--84. Springer Berlin Heidelberg (1993)

\bibitem{ShinAndo2011RoleBehaviorAF}
Ando, S., Suzuki, E.: Role-behavior analysis from trajectory data by cross-domain learning. In: Proceedings of the 11th IEEE International Conference on Data Mining. pp. 21--30 (2011)

\bibitem{StefanAtev2006LearningTP}
Atev, S., Masoud, O., Papanikolopoulos, N.: Learning traffic patterns at intersections by spectral clustering of motion trajectories. In: Proceedings of the IEEE/RSJ International Conference on Intelligent Robots and Systems. pp. 4851--4856 (2006)

\bibitem{borg2012applied}
Borg, I., Groenen, P.J., Mair, P.: Applied multidimensional scaling. Springer Science \& Business Media (2012)

\bibitem{sklearn_api}
Buitinck, L., Louppe, G., Blondel, M., Pedregosa, F., Mueller, A., Grisel, O., Niculae, V., Prettenhofer, P., Gramfort, A., Grobler, J., Layton, R., VanderPlas, J., Joly, A., Holt, B., Varoquaux, G.: {API} design for machine learning software: experiences from the scikit-learn project. In: Proceedings of the ECML PKDD Workshop: Languages for Data Mining and Machine Learning. pp. 108--122 (2013)

\bibitem{chen2018local}
Chen, B., Ting, K.M., Washio, T., Zhu, Y.: Local contrast as an effective means to robust clustering against varying densities. Machine Learning  \textbf{107}(8),  1621--1645 (2018)

\bibitem{chen2011clustering}
Chen, J., Wang, R., Liu, L., Song, J.: Clustering of trajectories based on {H}ausdorff distance. In: Proceedings of the International Conference on Electronics, Communications and Control. pp. 1940--1944. IEEE (2011)

\bibitem{KaiXuanChen2018RiemannianKB}
Chen, K.X., Wu, X.J., Wang, R., Kittler, J.: Riemannian kernel based nystr{\"o}m method for approximate infinite-dimensional covariance descriptors with application to image set classification. In: Proceedings of the 24th International Conference on Pattern Recognition. pp. 651--656. IEEE (2018)

\bibitem{chen2005robust}
Chen, L., {\"O}zsu, M.T., Oria, V.: Robust and fast similarity search for moving object trajectories. In: Proceedings of the ACM SIGMOD International Conference on Management of Data. pp. 491--502 (2005)

\bibitem{ZaibenChen2011DiscoveringPR}
Chen, Z., Shen, H.T., Zhou, X.: Discovering popular routes from trajectories. In: Proceedings of the 27th IEEE International Conference on Data Engineering. pp. 900--911 (2011)

\bibitem{CORDEIRODEAMORIM20121061}
{Cordeiro de Amorim}, R., Mirkin, B.: Minkowski metric, feature weighting and anomalous cluster initializing in k-means clustering. Pattern Recognition  \textbf{45}(3),  1061--1075 (2012)

\bibitem{cover1999elements}
Cover, T.M.: Elements of information theory. John Wiley \& Sons (1999)

\bibitem{eiter1994computing}
Eiter, T., Mannila, H.: Computing discrete {F}r{\'e}chet distance  (1994)

\bibitem{ester1996density}
Ester, M., Kriegel, H.P., Sander, J., Xu, X.: A density-based algorithm for discovering clusters in large spatial databases with noise. In: Proceedings of the Second International Conference on Knowledge Discovery and Data Mining. p. 226–231. AAAI Press (1996)

\bibitem{ZiquanFang2021E2D}
Fang, Z., Du, Y., Chen, L., Hu, Y., Gao, Y., Chen, G.: {E$^2$DTC}: An end to end deep trajectory clustering framework via self-training. In: Proceedings of the 37th IEEE International Conference on Data Engineering. pp. 696--707 (2021)

\bibitem{gao2020ship}
Gao, M., Shi, G.Y.: Ship-handling behavior pattern recognition using {AIS} sub-trajectory clustering analysis based on the t-{SNE} and spectral clustering algorithms. Ocean Engineering  \textbf{205},  106919 (2020)

\bibitem{ChihChiehHung2015ClusteringAA}
Hung, C.C., Peng, W.C., Lee, W.C.: Clustering and aggregating clues of trajectories for mining trajectory patterns and routes. The VLDB Journal  \textbf{24},  169--192 (2015)

\bibitem{keogh2000scaling}
Keogh, E.J., Pazzani, M.J.: Scaling up dynamic time warping for datamining applications. In: Proceedings of the Sixth ACM SIGKDD International Conference on Knowledge Discovery and Data Mining. pp. 285--289 (2000)

\bibitem{lee2007trajectory}
Lee, J.G., Han, J., Whang, K.Y.: Trajectory clustering: a partition-and-group framework. In: Proceedings of the ACM SIGMOD International Conference on Management of Data. pp. 593--604 (2007)

\bibitem{XiuchengLi2018DeepRL}
Li, X., Zhao, K., Cong, G., Jensen, C.S., Wei, W.: Deep representation learning for trajectory similarity computation. In: Proceedings of the 34th IEEE International Conference on Data Engineering. pp. 617--628. IEEE (2018)

\bibitem{liu2005motion}
Liu, C., Torralba, A., Freeman, W.T., Durand, F., Adelson, E.H.: Motion magnification. ACM Transactions on Graphics  \textbf{24}(3),  519--526 (2005)

\bibitem{KrikamolMuandet2016KernelME}
Muandet, K., Fukumizu, K., Sriperumbudur, B., Sch{\"o}lkopf, B., et~al.: Kernel mean embedding of distributions: A review and beyond. Foundations and Trends in Machine Learning  \textbf{10}(1-2),  1--141 (2017)

\bibitem{CMyers1980PerformanceTI}
Myers, C., Rabiner, L., Rosenberg, A.: Performance tradeoffs in dynamic time warping algorithms for isolated word recognition. IEEE Transactions on Acoustics, Speech, and Signal Processing  \textbf{28}(6),  623--635 (1980)

\bibitem{BoazNadler2006FundamentalLO}
Nadler, B., Galun, M.: Fundamental limitations of spectral clustering. Advances in Neural Information Processing Systems  \textbf{19} (2006)

\bibitem{rubner1998metric}
Rubner, Y., Tomasi, C., Guibas, L.J.: A metric for distributions with applications to image databases. In: Proceedings of the Sixth International Conference on Computer Vision. pp. 59--66 (1998)

\bibitem{TobiasSchreck2008VisualCA}
Schreck, T., Bernard, J., Von~Landesberger, T., Kohlhammer, J.: Visual cluster analysis of trajectory data with interactive kohonen maps. Information Visualization  \textbf{8}(1),  14--29 (2009)

\bibitem{shi1998motion}
Shi, J., Malik, J.: Motion segmentation and tracking using normalized cuts. In: Proceedings of the Sixth International Conference on Computer Vision. pp. 1154--1160. IEEE (1998)

\bibitem{XuanSong2016DeeptransportPA}
Song, X., Kanasugi, H., Shibasaki, R.: Deeptransport: Prediction and simulation of human mobility and transportation mode at a citywide level. In: Proceedings of the Twenty-Fifth International Joint Conference on Artificial Intelligence. pp. 2618--2624 (2016)

\bibitem{steinley2004properties}
Steinley, D.: Properties of the hubert-arable adjusted rand index. Psychological methods  \textbf{9}(3), ~386 (2004)

\bibitem{HanSu2020ASO}
Su, H., Liu, S., Zheng, B., Zhou, X., Zheng, K.: A survey of trajectory distance measures and performance evaluation. The VLDB Journal  \textbf{29},  3--32 (2019)

\bibitem{AbdelAzizTaha2015AnEA}
Taha, A.A., Hanbury, A.: An efficient algorithm for calculating the exact {H}ausdorff distance. IEEE Transactions on Pattern Analysis and Machine Intelligence  \textbf{37}(11),  2153--2163 (2015)

\bibitem{IDK-timeseries-VLDB2022}
Ting, K.M., Liu, Z., Zhang, H., Zhu, Y.: A new distributional treatment for time series and an anomaly detection investigation. Proceedings of the VLDB Endowment  \textbf{15}(11),  2321--2333 (2022)

\bibitem{KaiMingTing2020IsolationDK}
Ting, K.M., Xu, B.C., Washio, T., Zhou, Z.H.: Isolation distributional kernel: A new tool for point and group anomaly detections. IEEE Transactions on Knowledge and Data Engineering  \textbf{35}(03),  2697--2710 (2023)

\bibitem{KaiMingTing2018IsolationKA}
Ting, K.M., Zhu, Y., Zhou, Z.H.: Isolation kernel and its effect on svm. In: Proceedings of the 24th ACM SIGKDD International Conference on Knowledge Discovery and Data Mining. pp. 2329--2337 (2018)

\bibitem{wang2021survey}
Wang, S., Bao, Z., Culpepper, J.S., Cong, G.: A survey on trajectory data management, analytics, and learning. ACM Computing Surveys  \textbf{54}(2),  1--36 (2021)

\bibitem{ShengWang2019FastLT}
Wang, S., Bao, Z., Culpepper, J.S., Sellis, T., Qin, X.: Fast large-scale trajectory clustering. Proceedings of the VLDB Endowment  \textbf{13}(1),  29--42 (2019)

\bibitem{wang2023principled}
Wang, Y., Ting, K.M., Shang, Y.: A principled distributional approach to trajectory similarity measurement. arXiv preprint arXiv:2301.00393  (2023)

\bibitem{AJWard1954AGO}
Ward, A.: A generalization of the frechet distance of two curves. Proceedings of the National Academy of Sciences  \textbf{40}(7),  598--602 (1954)

\bibitem{xu2015unsupervised}
Xu, H., Zhou, Y., Lin, W., Zha, H.: Unsupervised trajectory clustering via adaptive multi-kernel-based shrinkage. In: Proceedings of the IEEE International Conference on Computer Vision. pp. 4328--4336 (2015)

\bibitem{yang2022fast}
Yang, W., Wang, S., Sun, Y., Peng, Z.: Fast dataset search with earth mover's distance. Proceedings of the VLDB Endowment  \textbf{15}(11),  2517--2529 (2022)

\bibitem{yao2019computing}
Yao, D., Cong, G., Zhang, C., Bi, J.: Computing trajectory similarity in linear time: A generic seed-guided neural metric learning approach. In: Proceedings of the 35th IEEE International Conference on Data Engineering. pp. 1358--1369. IEEE (2019)

\bibitem{yao2022trajgat}
Yao, D., Hu, H., Du, L., Cong, G., Han, S., Bi, J.: {TrajGAT}: A graph-based long-term dependency modeling approach for trajectory similarity computation. In: Proceedings of the 28th ACM SIGKDD Conference on Knowledge Discovery and Data Mining. pp. 2275--2285 (2022)

\bibitem{zhang2016red}
Zhang, Z., Huang, K., Tan, T., Yang, P., Li, J.: Red-sfa: Relation discovery based slow feature analysis for trajectory clustering. In: Proceedings of the IEEE Conference on Computer Vision and Pattern Recognition. pp. 752--760 (2016)

\bibitem{IDKC}
Zhu, Y., Ting, K.M.: Kernel-based clustering via isolation distributional kernel. Information Systems  \textbf{117},  102212 (Jul 2023)

\end{thebibliography}

\end{document}